%% file: main.tex
\documentclass{article}

\usepackage[preprint,nonatbib]{neurips_2024}

\usepackage[utf8]{inputenc} 
\usepackage[T1]{fontenc}    
\usepackage[pdftex]{graphicx}
\usepackage{hyperref}       
\usepackage{url}            
\usepackage{booktabs}       
\usepackage{amsfonts}       
\usepackage{nicefrac}       
\usepackage{microtype}      
\usepackage{xcolor}         

\usepackage{multirow}
\usepackage{biblatex}
\usepackage{pifont}
\usepackage{array}

\title{Integrating Spatial and Frequency Information \\ for Under-Display Camera Image Restoration}

\author{
\textbf{Kyusu Ahn}$^{1,3}$~~~
\textbf{Jinpyo Kim}$^{2}$~~~
\textbf{Chanwoo Park}$^{2}$~~~
\textbf{JiSoo Kim}$^{1}$~~~
\textbf{Jaejin Lee}$^{1,2}$~~~
\smallskip
\\
\small
$^1$Dept. of Data Science, Seoul National University, Seoul, Republic of Korea \\
\small
$^2$Dept. of Computer Science and Engineering, Seoul National University, Seoul, Republic of Korea \\
\small
$^3$Research Center, Samsung Display Co., Ltd., Yongin, Republic of Korea \\
\small
kyusu.ahn@snu.ac.kr, jinpyo@aces.snu.ac.kr, $\{$99chanwoo,jisoo.kim,jaejin$\}$@snu.ac.kr \\ 
\href{https://github.com/mcrl/SFIM}{https://github.com/mcrl/SFIM}
}

\begin{document}
\maketitle

\input{sec/00_Abstract}

\input{sec/01_Introduction}
\input{sec/02_Related_Work}
\input{sec/03_Architecture}
\input{sec/04_Experiments}
\input{sec/06_Conclusion}
\input{sec/07_Acknowledgment}

\printbibliography

\clearpage

\clearpage
\input{sec/09_Appendix}

\end{document}

%% file: sec/00_Abstract.tex
\begin{abstract}
Under-Display Camera (UDC) houses a digital camera lens under a display panel. However, UDC introduces complex degradations such as noise, blur, decrease in transmittance, and flare. Despite the remarkable progress, previous research on UDC mainly focuses on eliminating diffraction in the spatial domain and rarely explores its potential in the frequency domain. It is essential to consider both the spatial and frequency domains effectively. For example, degradations, such as noise and blur, can be addressed by local information (e.g., CNN kernels in the spatial domain). At the same time, tackling flares may require leveraging global information (e.g., the frequency domain). In this paper, we revisit the UDC degradations in the Fourier space and figure out intrinsic frequency priors that imply the presence of the flares. Based on this observation, we propose a novel multi-level DNN architecture called SFIM. It efficiently restores UDC-distorted images by integrating local and global (the collective contribution of all points in the image) information. The architecture exploits CNNs to capture local information and FFT-based models to capture global information. SFIM comprises a spatial domain block (SDB), a Frequency Domain Block (FDB), and an Attention-based Multi-level Integration Block (AMIB). Specifically, SDB focuses more on detailed textures such as noise and blur, FDB emphasizes irregular texture loss in extensive areas such as flare, and AMIB enables effective cross-domain interaction. SFIM's superior performance over state-of-the-art approaches is demonstrated through rigorous quantitative and qualitative assessments across three UDC benchmarks.

\end{abstract}

%% file: sec/01_Introduction.tex
\section{Introduction}\label{sec:intro}

Under-Display Camera (UDC) enables a full-screen display by housing a digital camera lens under the display panel. Modern smartphones, including the Samsung Galaxy Z-Fold series~\cite{samsungGalaxyZFold3,samsungGalaxyZFold4,samsungGalaxyZFold5} and the ZTE Axon series~\cite{zteAxon20,zteAxon30,zteAxon40-ultra} have adopted UDCs. Although UDC allows the complete removal of camera holes in the display panel, it severely degrades image quality. 

\begin{figure}
  \centering   
  \includegraphics[width=0.85\linewidth, bb=0 0 650 300]{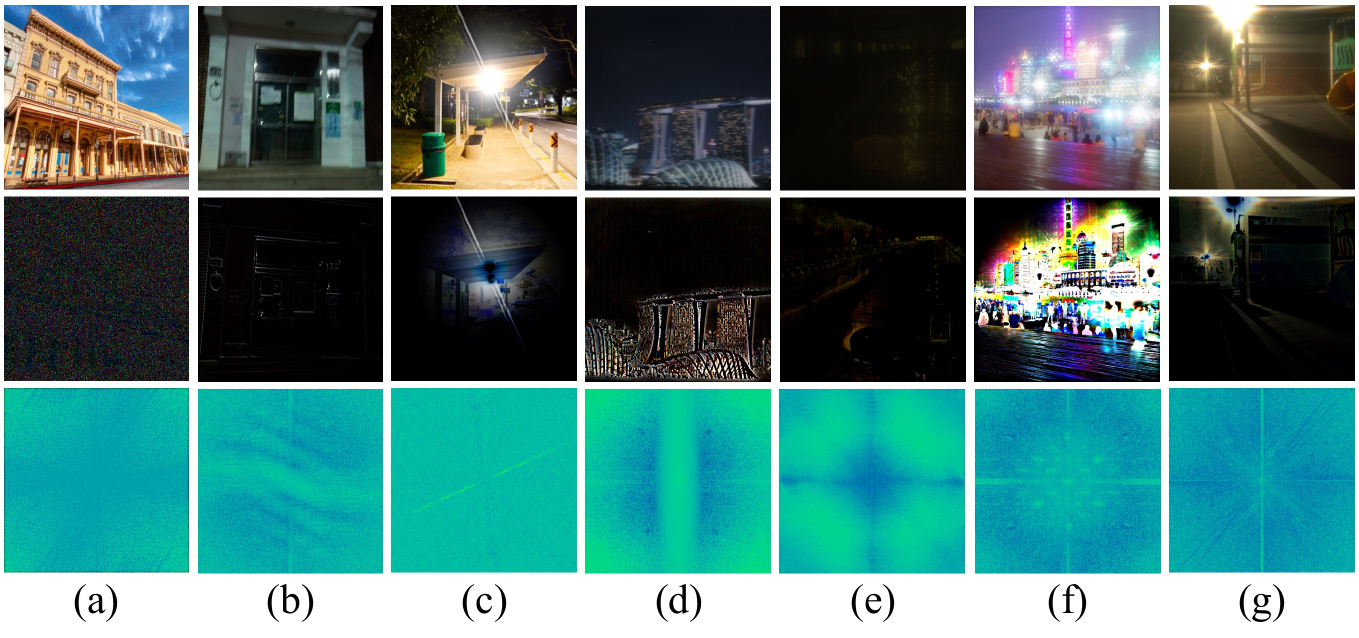}
   \caption{Spatial and frequency analysis for the various degradations, such as 
  (a) gaussian noise~\cite{huang2015single}, (b) blur~\cite{nah2017deep}, (c) lens flare~\cite{dai2022flare7k}, (d) T-OLED~\cite{zhou2021image}, (e) P-OLED~\cite{zhou2021image}, (f) SYNTH~\cite{feng2021removing}, and (g) UDC-SIT~\cite{ahn2024udc}. The existing UDC datasets span from (d) to (g).
  In the second row, pixel value differences between degraded and ground-truth images are represented in the spatial domain. In (a), pixel values are magnified ten times to improve noise visualization. The third row shows the frequency domain spectral amplitude differences between the degraded and ground-truth images.} 
   \label{fig:01_frequency_analysis}
\end{figure}

UDC degradations typically include low transmittance, blur, noise, and flares~\cite{ahn2024udc,feng2022mipi,kwon2021controllable,song2023under}. Restoring UDC images is more challenging than conventional degradations due to the simultaneous and severe occurrence of \textit{multiple degradation patterns across extensive areas} in a single UDC image. Conventional noise~\cite{huang2015single} and blur~\cite{nah2017deep} can be effectively eliminated using CNN in the spatial domain, as these degradations exhibit primarily local characteristics, as shown in Figure~\ref{fig:01_frequency_analysis}(a) and (b). However, addressing UDC degradations solely with local information is insufficient. 

Managing T-OLED/P-OLED datasets~\cite{zhou2021image} (Figure~\ref{fig:01_frequency_analysis}(d) and (e)) is relatively easy using local information. This is because they are captured with T-OLED and P-OLED panels positioned in front of a camera-based imaging system, hence lacking the distinctive flare characteristic of UDC. T-OLED dataset~\cite{zhou2021image} appears to resemble blurred images, and P-OLED dataset~\cite{zhou2021image} exhibits an excessive decrease in transmittance compared to the degradation of existing smartphones. 

On the other hand, SYNTH~\cite{feng2021removing} and UDC-SIT~\cite{ahn2024udc} datasets manifest flares that affect almost the entire image as depicted in Figure~\ref{fig:01_frequency_analysis}(f) and (g). SYNTH is produced by convolving a UDC's measured point spread function (PSF) with ground-truth images, and a real UDC smartphone captures UDC-SIT. Similar to the conventional lens flare~\cite{dai2022flare7k} shown in Figure~\ref{fig:01_frequency_analysis}(c), the difference in spectral amplitude between degraded and ground-truth images in SYNTH and UDC-SIT reveals distinct long, thin peaks, indicating the presence of flares in specific frequency components as depicted in Figure~\ref{fig:01_frequency_analysis}(f) and (g). Moreover, both SYNTH and UDC-SIT datasets have complex degradation patterns evident in both spatial and frequency difference maps as illustrated in Figure~\ref{fig:01_frequency_analysis}(f) and (g). Given UDC-SIT's UHD resolution, prioritizing global information is crucial for restoring flares across the entire image. 

Studies on UDC image restoration fall into two categories. One focuses on learning UDC degradation in the spatial domain. ECFNet~\cite{zhu2022enhanced} and UDC-UNet~\cite{liu2022udc} using CNNs with a multi-level architecture have achieved top rankings in the recent UDC MIPI challenge~\cite{feng2022mipi}. However, CNNs have fundamental problems when restoring UDC images. Liang et al.~\cite{liang2021swinir} describe that convolutions are ineffective for capturing long-range dependencies due to the local processing nature of CNNs. Thus, a multi-level CNN architecture to increase the receptive field is insufficient to tackle long-range dependencies. Another approach in this category employs Transformers for UDC image restoration~\cite{feng2022mipi}, but satisfactory results still need to be achieved. These methods, adopt spatial domain learning exclusively, overlooking diffraction characteristics (e.g., flares) in the frequency domain.

The other class of approaches explores the frequency domain. PDCRN~\cite{panikkasseril2020transform} uses discrete wavelet transform (DWT) for downsampling and upsampling, but the remaining operations are carried out using CNNs. FSI~\cite{liu2023fsi} uses a dual-stream network of spatial and frequency learning blocks. While showing improved performance compared to previous CNN-based models focusing solely on the spatial domain, they still face challenges in recovering irregular texture loss in extensive areas.

In this paper, we propose SFIM (\textbf{S}patial and \textbf{F}requency \textbf{I}nteractive learning in a \textbf{M}ulti-level architecture) for UDC image restoration. The critical insight of SFIM is to harness the inherent capability of CNNs, FFT-based models, and multi-level architecture. CNNs, using kernels, excel at capturing local information. On the other hand, using FFT-based models enables the extraction of global information since a frequency component is the collective contribution of all points in the image. In SFIM, CNNs operate at the upper level with a narrow receptive field to leverage noise and blur, while FFT-based models operate at the lower level with a wider receptive field to handle flares.

The contributions of this paper are summarized as follows:
\begin{itemize}

\item We propose spatial and frequency interactive learning in a multi-level architecture (SFIM), using CNNs (at level 1) and FFT-based models (at lower levels). We verify that SFIM successfully eliminates UDC flares more effectively than the previous state-of-the-art models.

\item We propose an attention-based multi-level integration (AMI) technique, which effectively integrates all levels of the multi-level architecture and guides the network to attend to features associated with flares.

\item Our experiments demonstrate that SFIM significantly outperforms existing state-of-the-art models in UDC benchmarks that contain UDC flares. Moreover, SFIM shows the best qualitative performance on irregular texture loss in extensive areas. 
\end{itemize}

%% file: sec/02_Related_Work.tex
\section{Related Work}

\paragraph{Low-level image restoration tasks.} 
Low-level image restoration tasks focuses on intrinsic characteristics to improve the overall quality and clarity. They include image denoising~\cite{wang2022uformer, zamir2020cycleisp}, deblurring~\cite{abuolaim2020defocus,kong2023efficient, zamir2022restormer}, deraining~\cite{jiang2020multi,wang2019spatial}, lens flare removal~\cite{dai2022flare7k,zhou2023improving}, edge detection~\cite{poma2020dense}, and color correction~\cite{xu2019towards}.
Prior studies primarily focus on an individual task, but the UDC images require a complex combination of multiple low-level tasks. This paper focuses on a combination of low-level restoration tasks for UDC images, particularly addressing flares that differ from lens flares in their causes and patterns. 

\paragraph{UDC datasets.}
There are many UDC datasets to train and evaluate image restoration models. Zhou et al.~\cite{zhou2021image} build a dataset using a monitor-based imaging system (T-OLED/P-OLED). 
Feng et al.~\cite{feng2021removing} present a synthetic dataset by convolving the measured PSF of a UDC with ground-truth images (SYNTH). 
Feng et al.~\cite{feng2023generating} also create a pseudo-real UDC dataset using two cameras. They use AlignFormer~\cite{feng2023generating} to align the images from the two cameras. 
Ahn et al.~\cite{ahn2024udc} propose a real-world UDC dataset named UDC-SIT of UHD resolution captured by a real UDC smartphone, effectively reflecting real-world UDC degradation with high alignment accuracy.   

\paragraph{UDC image restoration models.}
The majority of the UDC image restoration studies are based on CNN architectures. For instance, Zhu et al. ~\cite{zhu2022enhanced} propose ECFNet to incorporate information across multiple scales of images. Liu et al.~\cite{liu2022udc} propose UDC-UNet, which has condition branches for spatially variant manipulation and kernel branches to incorporate prior knowledge (i.e., the PSFs).
In UDC image restoration challenges~\cite{feng2022mipi, zhou2020udc}, there exist approaches to use SwinTransfomer~\cite{liu2021swin}. However, the quality of the restored images from SwinTransformer-based models is not satisfactory enough compared to the CNN-based models.
There is also an attempt~\cite{panikkasseril2020transform} to use DWT for downsampling and upsampling, but the majority of the operations are conducted using CNNs. Moreover, their model does not address flare removal because flares are absent in T-OLED and P-OLED datasets they use. FSI~\cite{liu2023fsi} considers both spatial and frequency domains using a dual-stream network. However, irregular texture loss in extensive areas still presents challenges.

%% file: sec/03_Architecture.tex
\section{Architecture of SFIM}
\label{sec:03_SFIM}
\begin{figure}
  \centering  
  \includegraphics[width=0.9\textwidth, bb=0 0 600 350]{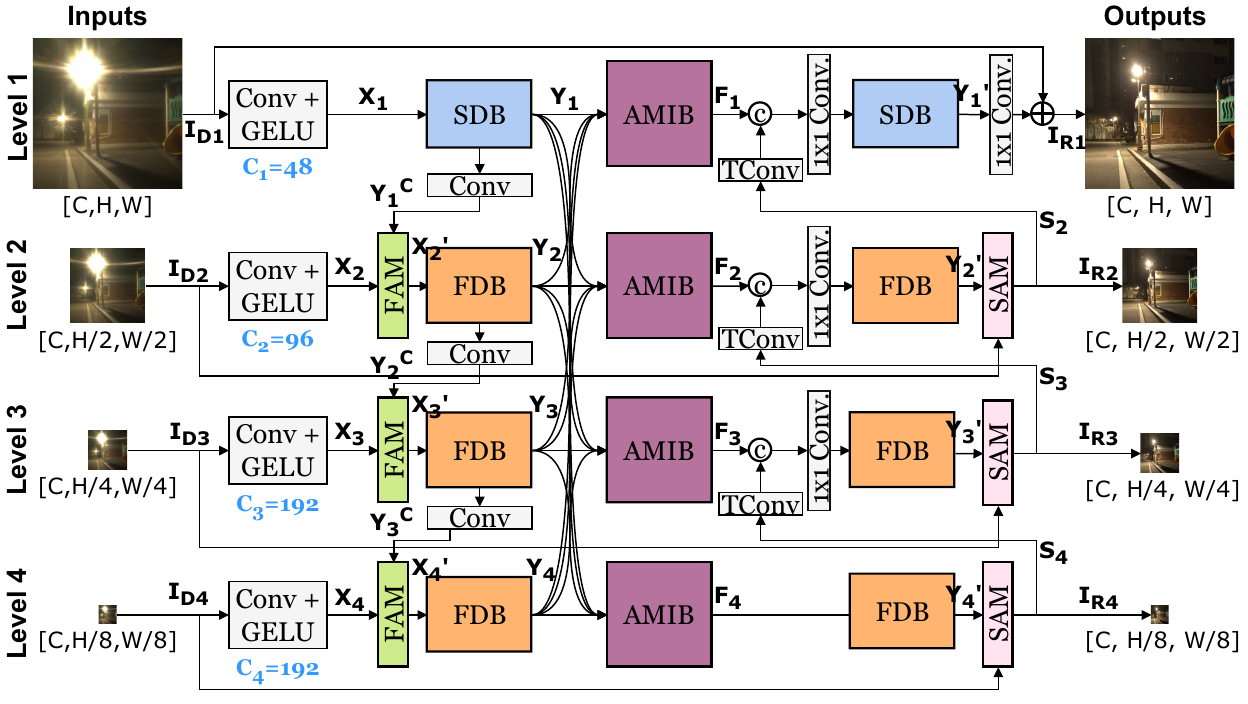}
  \caption{Overall architecture of SFIM.}
  \label{fig:03_SFIM}
\end{figure}

In this section, we describe the SFIM architecture and the rationale behind the design. It is mainly composed of spatial domain block (SDB), frequency domain block (FDB), and attention-based multi-level integration block (AMIB) as shown in Figure~\ref{fig:03_SFIM}. 

\subsection{Overview}
\label{sec:03_Motivation}
Our goal is to present an effective and efficient method to explore the UDC's \textit{multiple degradation patterns across extensive areas} as shown in Figure~\ref{fig:01_frequency_analysis}(f) and (g). The UDC exhibits locally manageable degradations, such as noise and blur, while flare presents a challenge when relying solely on local information. To address this, we consider the capability of a multi-level architecture that consists of CNN and FFT operations. 

In the multi-level structure, downsampling from higher to lower levels leads to a loss of high-frequency information and an increase in the receptive field. Consequently, the upper level, enriched with high-frequency components, suits CNNs' proficiency in \textit{local feature extraction} using a kernel (e.g., restoring sharp edges from blurred or noisy images). However, CNNs struggle to capture long-range dependencies~\cite{liu2022udc}, and even at lower levels, they lack sufficient receptive field, especially for leveraging flares of the UDC-SIT~\cite{ahn2024udc} dataset with UHD resolution. Incorporating FFT-based models compensates for this limitation. Please see more in Appendix~\ref{sec:appendix_background}.

While FFT operations may face challenges restoring sharp edges~\cite{koh2022bnudc,koh2021single}, they excel in handling global features like flares. Moreover, the Fourier space enables figuring out \textit{intrinsic frequency priors} that imply the presence of the flares. Note that multi-level architectures using CNNs at all levels, such as UDCUNet~\cite{liu2022udc} and ECFNet~\cite{zhu2022enhanced} or using FFT operations at every level, such as FFTformer~\cite{kong2023efficient}, perform worse than SFIM, particularly on the UDC-SIT dataset (Table~\ref{tab:exp_quantitative_sfim}). The deliberate arrangement of components in the SFIM architecture optimally harnesses the unique strength of each component.

Since SFIM adopts a four-level architecture with a coarse-to-fine-grained approach~\cite{cho2021rethinking}, inputs of various dimensions are introduced at each level. The input to the SFIM is the UDC-degraded image $I_{D_i} \in \mathbb{R}^{C \times \frac{H}{2^{i-1}} \times \frac{W}{2^{i-1}}}$, and the output is the restored image $I_{R_i} \in \mathbb{R}^{C \times \frac{H}{2^{i-1}} \times \frac{W}{2^{i-1}}}$. The input images undergo a convolution and GELU activation~\cite{hendrycks2016gaussian} and become a shallow feature $X_i \in \mathbb{R}^{C_i \times \frac{H}{2^{i-1}} \times \frac{W}{2^{i-1}}}$ ($C_i \in \{48,96,192\}$). 

At each level, features are encoded or decoded by SDBs or FDBs in the spatial or frequency domain. They are integrated by the feature attention modules (FAMs), the supervised attention modules (SAMs), and the proposed AMIBs.
    
\subsection{Spatial Domain Blocks}
\label{sec:03_SDB}
\begin{figure}
  \centering
  \begin{small}
    \begin{minipage}[t]{0.45\linewidth}
        \centering
        \includegraphics[width=\linewidth]{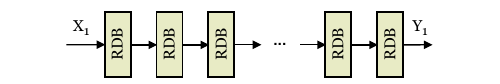}
        (a) Spatial domain block (SDB)
    \end{minipage}
    \begin{minipage}[t]{0.50\linewidth}
        \centering
        \includegraphics[width=\linewidth]{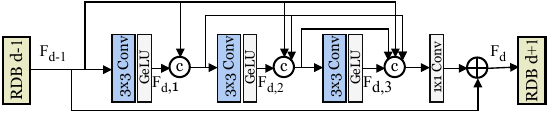}
        (b) Residual dense block (RDB)
    \end{minipage}
  \end{small}   
  \caption{The structure of the SDB that consists of eight RDBs.}
  \label{fig:03_SDB}

  \vspace{1.0\baselineskip}
  \begin{minipage}{\linewidth}
  \centering
  \includegraphics[width=1.0\textwidth]{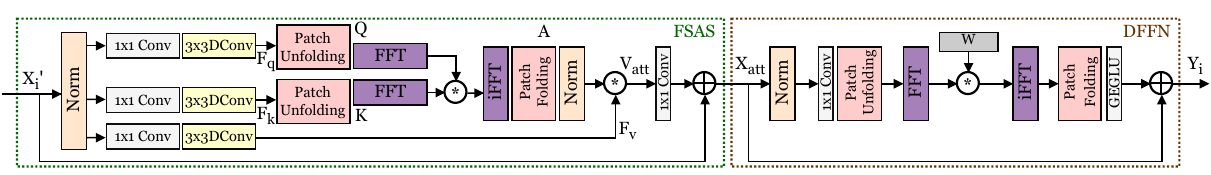}
  \caption{The structure of the FDB that consists of an FSAS and a DFFN.}
  \label{fig:03_FDB}
  \end{minipage}
\end{figure}

SDBs mainly remove the blur and noise at level 1, as described in Figure~\ref{fig:03_SFIM}. They are used at level 1 because layers in the first level see the image's most compact areas compared to the other three levels. Unlike levels 2, 3, and 4, all operations at level 1 are performed in the spatial domain. SDB consists of eight residual dense blocks (RDBs)~\cite{zhang2018residual} (Figure~\ref{fig:03_SDB}(a)). An RDB uses all the features within it via local dense connections (Figure~\ref{fig:03_SDB}(b)). The output of the $c^{th}$ convolutional layer of the $d^{th}$ RDB is as follows:

\begin{equation}
  \label{eq:rdb}
    F_{d,c} = \mathcal{G}(W_{d,c}([F_{d-1}, F_{d,1}, ..., F_{d,c-1}])),
\end{equation}

where $\mathcal{G}$ denotes the GELU activation, and $W_{d,c}$ represents the weights of the $c^{th}$ convolutional layer ($c \in \{1,2,3\}$) in the $d^{th}$ RDB ($d \in \{1,2, ..., 8\}$). The bias term is excluded for simplicity. The notation $[F_{d-1}, F_{d,1}, ..., F_{d,c-1}]$ denotes the concatenation of features generated by the $(d-1)^{th}$ RDB and the convolutional layers from 1 to $c-1$ in the $d^{th}$ RDB. This results in $G_0 + (c-1) \times G$ features, where $F_{d,c}$ consists of G features. The $1 \times 1$ convolution adaptively preserves the accumulated features. Thus, an SDB takes $X_1 \in \mathbb{R}^{C_1 \times H \times W}$ as input and outputs $Y_1 \in \mathbb{R}^{C_1 \times H \times W}$.

\subsection{Frequency Domain Blocks} 
FDBs are mainly used to mitigate the flares from level 2 to level 4, as illustrated in Figure~\ref{fig:03_SFIM}. FFT-based models can capture global information since a frequency component is a collective contribution of all points. Moreover, while convolution is spatially invariant, FFT-based models can effectively address missing spatial variations, such as spatially variant flares in the UDC-SIT dataset. 

An FDB consists of a frequency domain-based self-attention solver (FSAS) and discriminative frequency domain-based feed-forward network (DFFN)~\cite{kong2023efficient} as shown in Figure~\ref{fig:03_FDB}. The FDB takes $X_i^{\prime} \in \mathbb{R}^{C_i \times \frac{H}{2^{i-1}} \times \frac{W}{2^{i-1}}}$ as input and outputs $Y_i \in \mathbb{R}^{C_i \times \frac{H}{2^{i-1}} \times \frac{W}{2^{i-1}}}$ (for levels $i \in \{2,3,4\}$).

The FSAS is based on the convolution theorem that the correlation or convolution of two signals in the spatial domain is identical to their element-wise product in the frequency domain. For input feature $X_i^{\prime}$, layer normalization, $1 \times 1$ convolution, and $3 \times 3$ depth-wise convolution generate $F_q$, $F_k$, and $F_v$. Subsequently, patch unfolding $\mathcal{P}(\cdot)$ is applied to $F_q$ and $F_k$ to reduce the computational complexity of FFT, resulting in $Q$ and $K$. Then, FFT is employed on $Q$ and $K$, followed by an elementwise product between them, the inverse FFT, and patch folding $\mathcal{P}^{-1}(\cdot)$:
\begin{equation}
\label{eq:fsas_qkv}
\begin{array}{c}
F_q, F_k, F_v = DConv_{3\times3}(Conv_{1\times1}(\mathcal{N}(X_i^{\prime}))), \\
Q = \mathcal{P}(F_q) \quad \mbox{and} \quad K = \mathcal{P}(F_k), \\
A = \mathcal{P}^{-1}(\mathcal{F}^{-1}(\mathcal{F}(Q)\overline{\mathcal{F}(K)})),
\end{array}
\end{equation}
where $\mathcal{N}$ is a layer normalization, $DConv$ refers to a depthwise convolution, $\mathcal{F}(\cdot)$ is the FFT operation, $\mathcal{F}^{-1}(\cdot)$ is its inverse, and $\overline{\mathcal{F}(\cdot)}$ is the conjugate transpose operation. Then, the aggregated feature $V_{att}$ and output feature $X_{att}$ are acquired as follows:

\begin{equation}
  \label{eq:fsas_v}
  V_{att} = \mathcal{N}(A)F_v \quad \mbox{and} \quad X_{att} = X_i^{\prime} + Conv_{1 \times 1}(V_{att}).
\end{equation}
    
A DFFN adaptively preserves frequency information or \textit{intrinsic frequency priors} (e.g., flares), determining the importance of low and high frequencies. After applying layer normalization to $X_{att}$, a sequence of operations, including $1 \times 1$ convolution, patch unfolding, and FFT, is performed. Subsequently, an elementwise multiplication is performed with a learnable frequency weight matrix $\textbf{W} \in \mathbb{R}^{8 \times 8}$ that can decide essential frequency components. The process finishes with inverse FFT, patch folding, and GEGLU function $\mathcal{G}$~\cite{shazeer2020glu}. The output feature of the DFFN, which is also the output of an FDB, denoted as $Y_i$, is defined as follows:

\begin{equation}
\label{eq:dffn}
\begin{array}{c}
Z_1 = Conv_{1 \times 1}(\mathcal{N} (X_{att})), \quad
Z_{1}^{f} = \mathcal{F}(\mathcal{P}(Z_1)), \\
Z_2 = \mathcal{F}^{-1}(\textbf{W}Z_{1}^{f}), \quad \mbox{and} \quad
Y_{i} = \mathcal{G}(\mathcal{P}^{-1}(Z_2))+X_{att}.
\end{array}
\end{equation}

\subsection{Integrating Spatial-Frequency Information}
\label{sec:03_Integrating_SF}
\begin{figure}[t]
  \centering
  \begin{small}
    \begin{minipage}[t]{0.55\linewidth}
      \centering
      \includegraphics[width=\linewidth]{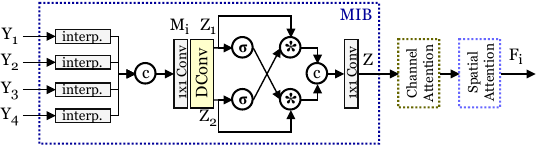}
      (a) AMIB
    \end{minipage}
    \begin{minipage}[t]{0.25\linewidth}
      \centering
      \includegraphics[width=\linewidth]{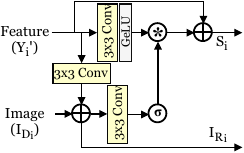}
      (b) SAM 
    \end{minipage}
    \begin{minipage}[t]{0.15\linewidth}
      \centering
      \includegraphics[width=\linewidth]{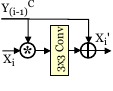}
      (c) FAM
    \end{minipage}
  \end{small}
  \caption{The structure of spatial and frequency integration modules. (a) The proposed attention-based multi-level integration block (AMIB). (b) SAM. (c) FAM.} 
  \label{fig:03_AMIB_FAM_SAM}
\end{figure}

We use FAM~\cite{cho2021rethinking} and SAM~\cite{zamir2021multi} to integrate information across levels. In addition to FAM and SAM, we newly introduce an attention-based multi-level integration block (AMIB). It seamlessly merges all four levels, directing the network to focus on crucial channel and spatial information. Figure~\ref{fig:03_AMIB_FAM_SAM} shows the structure of AMIB. 

Inspired by Cho et al.~\cite{cho2021rethinking}, the AMIB is mainly composed of a multi-level integration block (MIB), channel attention (CA), and spatial attention (SA). MIB receives features from all four levels. As these features have various dimensions, they are adjusted to achieve uniform dimensions, concatenated, and then undergo $1 \times 1$ convolution to match the number of channels to $Y_i$. This procedure ensures that each level incorporates both spatial and frequency information from all levels. For example, the fused feature $M_2$ at level 2 is as follows:

        \begin{equation}
          \label{eq:amib_concat}
          M_2 = Conv_{1 \times 1}([\mathcal{I}(Y_1), Y_2, \mathcal{I}(Y_3), \mathcal{I}(Y_4)]),
        \end{equation}
        
where $[\cdot]$ denotes the concatenation, $\mathcal{I}(\cdot)$ is an interpolation, $Y_1$ is the encoded feature by SDB at level 1, and $Y_2$, $Y_3$ and $Y_4$ are the encoded features by FDB at levels 2, 3 and 4, respectively.

The fused features are split into two features by depthwise convolution. After sigmoid activation and elementwise multiplication, they are concatenated again. The subsequent steps involve $1 \times 1$ convolution, channel attention, and spatial attention. These operations are described as follows:

\begin{equation}
\label{eq:amib_attn}
            Z_1, Z_2 = DConv_{3 \times 3}(M_i), \;
            Z = Conv_{1 \times 1}([\sigma(Z1)*Z2, \; \sigma(Z2)*Z1]), \; \mbox{and} \;
            F_{i} = \mathcal{SA}(\mathcal{CA}(Z))
\end{equation}

where $Dconv$, $\sigma(\cdot)$, $*$, $[\cdot]$, $\mathcal{CA}(\cdot)$, and $\mathcal{SA}(\cdot)$ denote the depthwise convolution, sigmoid function, elementwise multiplication, concatenation, channel attention, and spatial attention, respectively. 
 
The crucial channel is extracted from features across all levels through channel attention, and spatial attention accentuates important regions (e.g., flares). They enhance the model's ability to capture important details. We use both max pooling and average pooling following the approach outlined in CBAM~\cite{woo2018cbam} for channel and spatial attention. 
        
SAM shown in Figure~\ref{fig:03_AMIB_FAM_SAM}(b), produces two outputs: an attention map $S_i$ that enhances or suppresses features and a small restored image $I_{R_i}$ (outputs of levels 2, 3, and 4) that undergoes the backpropagation process. It plays a crucial role in progressive image restoration. For instance, the attention map at level 2 supports the feature learning process at level 1. Hence, the learned features by the FDB at level 2 affect the learned features by the SDB at level 1.

FAM shown in Figure~\ref{fig:03_AMIB_FAM_SAM}(c) computes the element-wise product of two input feature maps from the current (e.g., level 2; $X_2$) and the previous level (e.g., level 1; ${Y_1}^C$). Then, it applies a convolution and adds the feature map of the previous level (e.g., level 1; ${Y_1}^C$) to the result. SFIM uses FAM to emphasize the critical features from the previous level. Hence, the spatially learned features by the SDB at level 1 are transferred to the lower levels' learned features by the FDB.

\subsection{Loss Functions}
\label{sec:login_loss}
    In the multi-level DNN models, each level's output mimics the ground-truth image with an appropriate scale~\cite{cho2021rethinking,nah2017deep,zhu2022enhanced}. Following this approach, we employ a multi-scale loss function:
    
    \begin{equation}
      \label{eq:coarse_to_fine}
      \mathcal{L} = \sum\limits_{l=1}^{L} \mathcal{L}_l,
    \end{equation}
    
where $L$ is the number of levels and $\mathcal{L}_l$ is the loss at level $l$. We use the Charbonnier loss to measure the similarity of the restored image with the ground truth. The Charbonnier loss at level $l$ is:

    \begin{equation}
      \label{eq:charbonnier}
      \mathcal{L}_{Char, l} = \sqrt{|| R_l - G_l ||^2 + \epsilon^2},
    \end{equation}
    
where $R_l$ is the restored image at level $l$, $G_l$ is the ground truth image at level $l$, and $\epsilon$ is a constant set to $10^{-3}$. We also use the SSIM loss to measure the structural similarity between the restored image and the ground truth. The SSIM loss at level $l$ is defined as:

    \begin{equation}
      \label{eq:ssim}
      \mathcal{L}_{SSIM, l} = 1 - SSIM(R_l, G_l).
    \end{equation}
    
We use the FFT loss to measure the restored image quality in the frequency domain, which is useful in measuring the degradation in UDC images~\cite{ahn2024udc,zhu2022enhanced}. To calculate the FFT loss, we first apply FFT to the restored and ground truth images to obtain their \textit{amplitudes} ($|\mathcal{F}_{R_l}|$ and $|\mathcal{F}_{G_l}|$) and \textit{phases} ($\phi_{R_l}$ and $\phi_{G_l}$). We then take the L1 distance to quantify the differences in amplitude and phase. Thus, FFT loss of level $l$ has two terms: the amplitude term, $\mathcal{L}_{\mathcal{A}, l}$ and the phase term $\mathcal{L}_{\phi,l}$.

    \begin{equation}
      \label{eq:fft_amp}
      \mathcal{L}_{\mathcal{A}, l} = \sum\limits_{u=0}^{M_l-1}\sum\limits_{v=0}^{N_l-1} \big[ | \mathcal{F}_{R_l}(u,v) | - | \mathcal{F}_{G_l}(u,v) | \big] \quad \mbox{and} \quad
      \mathcal{L}_{\phi,l} = \sum\limits_{u=0}^{M_l-1}\sum\limits_{v=0}^{N_l-1} \big[\phi_{R_l} (u,v) - \phi_{G_l} (u,v) \big],
    \end{equation}
    
where $(u,v)$ is a point in the frequency domain, and $(M_l, N_l)$ is size of the image at level $l$. 
Finally, the loss at level $l$ is the weighted sum of $\mathcal{L}_{Char, l}$, $\mathcal{L}_{SSIM, l}$, $\mathcal{L}_{\mathcal{A}, l}$, and $\mathcal{L}_{\mathcal{\phi}, l}$.

    \begin{equation}
      \label{eq:loss_total}
      \mathcal{L}_{l} = \mathcal{L}_{Char, l} + \lambda_1 \cdot \mathcal{L}_{SSIM, l} + \lambda_2 \cdot \mathcal{L}_{\mathcal{A}, l} + \lambda_3 \cdot \mathcal{L}_{\phi, l}.
    \end{equation}
    
where $\lambda_1$, $\lambda_2$, and $\lambda_3$ are the weights. 

%% file: sec/04_Experiments.tex
\section{Experiments}
\label{sec:experiments}
In this section, we evaluate the performance of SFIM by comparing eight existing methods using three benchmark datasets.
\subsection{Experimental Settings}
\label{sec:datasets_and_metrics}
Among publicly available UDC datasets, we use T-OLED~\cite{zhou2021image}, SYNTH~\cite{feng2021removing}, and UDC-SIT~\cite{ahn2024udc} in our experiment. Notably, only SYNTH and UDC-SIT incorporate flares. UDC-SIT, with its UHD resolution, emphasizes integrating spatial and frequency domains (please see Appendix~\mbox{\ref{sec:appendix_background}}). This choice enables us to assess SFIM's performance on both synthetic and real-world datasets.

We evaluate the restored images using three quantitative metrics: PSNR, SSIM, and LPIPS. The LPIPS metric is applied only to 3-channel images (e.g., SYNTH). 

We use the AdamW optimizer~\cite{loshchilov2017decoupled} with $\beta_1=0.9, \beta_2=0.999, \lambda=0.01$. We empirically set the weights to $\lambda_1 = \lambda_2 = \lambda_3 = 1$. We use progressive learning strategy~\cite{zamir2022restormer,zhu2022enhanced}. Thus, SFIM is trained on smaller images initially, then gradually transitioning to larger images in later training phases. The detailed configuration is described in Appendix~\ref{sec:appendix_training_details}. 

\subsection{Comparison with Other Models}
\label{sec:04_comparison}
We compare SFIM with several state-of-the-art methods. To ensure fairness, we acquire performance data from the original paper or reproduce the results using the officially released models.

\paragraph{Quantitative comparison.}
\label{sec:04_quantitative_comparison}

Table~\ref{tab:exp_quantitative_sfim} compares quantitative performance between the methods. SFIM excels across all metrics compared to other models on UDC-SIT dataset. Notably, SFIM outperforms the state-of-the-art model by a significant margin, achieving a PSNR improvement of \textbf{1.35 dB}. Since FSI~\cite{liu2023fsi} cannot be applied to the 4-channel UDC-SIT dataset because its color correction module is designed for 3-channel images. Therefore, we convert the 4-channel UDC-SIT images to 3-channel images for FSI. For SYNTH dataset, SFIM outperforms the state-of-the-art model by a substantial margin and achieves a PSNR improvement of \textbf{0.40 dB}. SFIM surpasses multi-level architectures using CNNs for all levels (ECFNet and UDC-UNet) and those using FFT operations for all levels (FFTformer). For T-OLED dataset, please see Appendix~\ref{sec:appendix_quantitative}.

\begin{table}[b]
    \centering
      \caption{Restoration performance comparison. The LPIPS applies only to 3-channel images (e.g., SYNTH). ECFNet uses different embedding dimensions for different datasets, resulting in varying parameter counts. FSI is reevaluated with the official pre-trained model for the SYNTH dataset.} 
      \label{tab:exp_quantitative_sfim}
      \centering
      \resizebox{0.9\linewidth}{!}{
      \begin{tabular}{l|c|cc|ccc}
        \toprule
             & Param & \multicolumn{2}{c|}{UDC-SIT} & \multicolumn{3}{c}{SYNTH} \\
             & (M) & PSNR~$\uparrow$ & SSIM~$\uparrow$ & PSNR~$\uparrow$ & SSIM~$\uparrow$ & LPIPS~$\downarrow$ \\ \midrule
        Input       & -     & 21.06 & 0.7263 & 25.95 & 0.8568 & 0.2545 \\
        DISCNet~\cite{feng2021removing}     & 3.80  & 26.32 & 0.8457 & 43.06 & 0.9870 & 0.0113 \\
        UDC-UNet~\cite{liu2022udc}    & 14.00 & 27.44 & \underline{0.8637} & \underline{49.37} & 0.9933 & \underline{0.0065} \\
        Uformer-T~\cite{wang2022uformer}   & 5.23  & 27.28 & 0.8594 & 42.47 & 0.9844 & 0.0207 \\
        ECFNet~\cite{zhu2022enhanced}      & 38.72/16.85 & \underline{28.26} & 0.8491 & 47.75 & 0.9915 & 0.0085 \\
        SRGAN~\cite{ledig2017photo}       & 0.45 & 24.73 & 0.8139 & 33.13 & 0.9536 & 0.0568 \\
        SwinIR~\cite{liang2021swinir}      & 11.80 & 26.03 & 0.8460 & 39.06 & 0.9772 & 0.0273 \\
        FFTformer~\cite{kong2023efficient}   & 16.60 & 26.96 & 0.8606 & 42.30 & 0.9870 & 0.0131 \\
        FSI~\cite{liu2023fsi}         & 5.19  & 21.72 & 0.6674 & 46.14 & \underline{0.9932} & 0.0121 \\
        SFIM (ours) & 24.89 & \textbf{29.61} & \textbf{0.8758} & \textbf{49.77} & \textbf{0.9940} & \textbf{0.0053} \\
        \bottomrule
      \end{tabular}
      }
\end{table}

\begin{figure}[!b]
    \centering
    \begin{minipage}{\linewidth}
        \centering
        \includegraphics[width=0.98\textwidth]{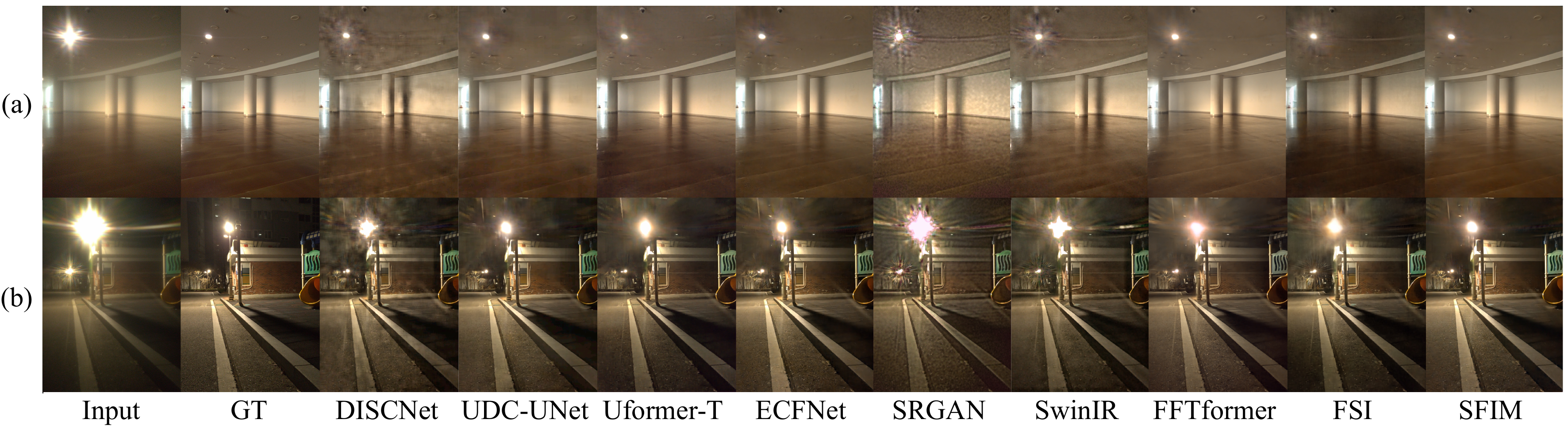}
        \caption{UDC-SIT~\cite{ahn2024udc} images restored by the models.}
        \label{fig:04_qualitative_results_udcsit}
    \end{minipage} \\
    \vspace{1.0\baselineskip}
    \begin{minipage}{\linewidth}
        \centering
        \includegraphics[width=1.0\textwidth]{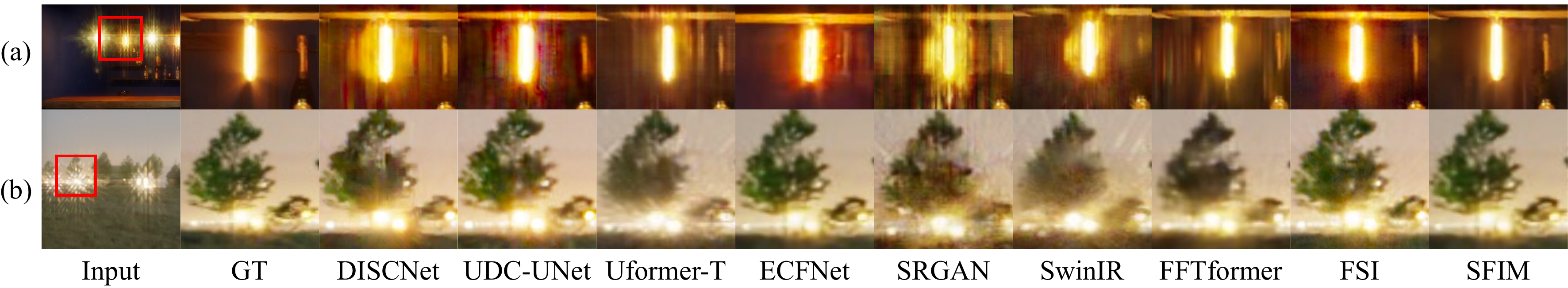}
        \caption{SYNTH~\cite{feng2021removing} images restored by the models.}
        \label{fig:04_qualitative_results_fengs}
    \end{minipage}
\end{figure}

\paragraph{Qualitative comparison.}
\label{sec:04_qualitative_comparison}
SFIM also improves visual quality, particularly addressing flare and texture loss caused by diffraction. In the UDC-SIT dataset, Figure~\ref{fig:04_qualitative_results_udcsit} shows that SFIM outperforms the others in eliminating the flare caused by distortion around artificial light sources. In the SYNTH dataset, the image restored by SFIM, as shown in Figure~\ref{fig:04_qualitative_results_fengs}(a), exhibits minimal distortion around the light source, and the restoration accurately captures details of the bottle adjacent to the light source. Moreover, SFIM excels in restoring intricate details around the tree, as depicted in Figure~\ref{fig:04_qualitative_results_fengs}(b). Please see additional samples in Appendix~\ref{sec:appendix_qualitative}.

\begin{table}[tb]
\centering
\caption{The effect of AMIB components on the UDC-SIT~\cite{ahn2024udc} dataset. The performance presented in this table corresponds to the SFIM model trained up to phase 1. Our proposed AMIB can be interpreted as "Base+MIB+CA+SA". \label{tab:individual_components}} 
\resizebox{0.5\columnwidth}{!}{
\begin{tabular}{cccc|cccc}
\toprule
Base      & MIB       & CA        & SA        & PSNR~$\uparrow$ & SSIM~$\uparrow$ \\ \midrule
\ding{52} &           &           &           & 28.19 & 0.8634 \\
\ding{52} & \ding{52} &           &           & 28.67 & 0.8704 \\
\ding{52} & \ding{52} & \ding{52} &           & 28.97 & 0.8678 \\
\ding{52} &           & \ding{52} & \ding{52} & 28.76 & 0.8648\\
\ding{52} & \ding{52} & \ding{52} & \ding{52} & \textbf{29.47} & \textbf{0.8742} \\ \bottomrule
\end{tabular}
}
\end{table}

\subsection{Ablation Study}
\label{sec:exp_ablation}
We conduct ablation study on each component of the proposed AMIB and explore the optimal configuration, embedding dimension, and number of levels in the SFIM architecture. The ablation study on FFT loss terms is presented in Appendix~\ref{sec:appendix_ablation}.
    
\begin{figure}[!t]
  \centering
  \begin{minipage}{\linewidth}
    \centering
    \includegraphics[width=0.52\linewidth]{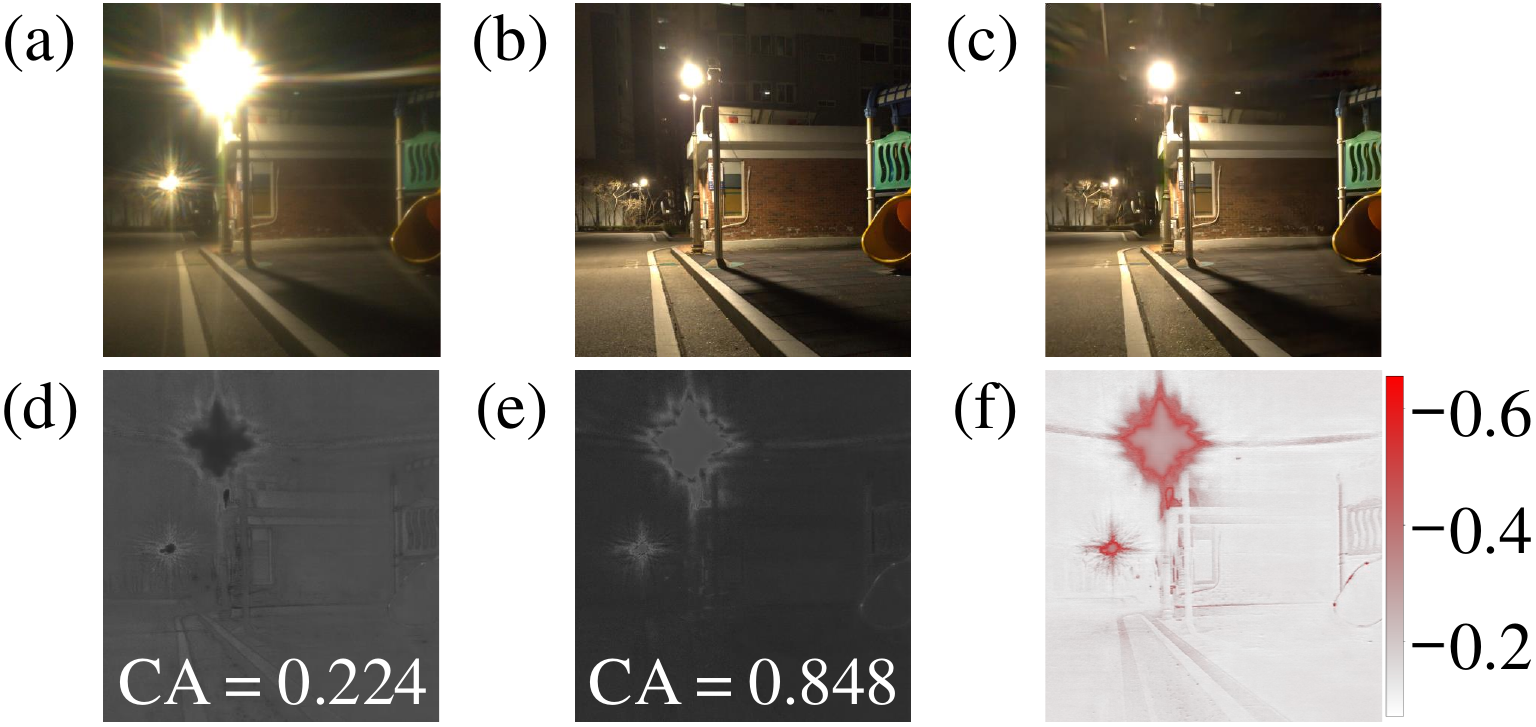}
    \vspace{-0.5\baselineskip}
    \caption{Attention map highlights flare regions. (a) Degraded. (b) GT. (c) Restored by SFIM. (d) A feature with a low CA score (0.224), and (e) a high CA score (0.848). (f) The spatial attention map.}
    \label{fig:amib_attn}
  \end{minipage}
  \vspace{-1.0\baselineskip}
\end{figure}

\paragraph{Individual components in AMIB.}
\label{sec:04_ablation_indivisual_components}
We use SFIM without the AMIB as the "Base" model and progressively add each AMIB component (i.e., MIB, CA, and SA) for comparison. Table~\ref{tab:individual_components} shows the impact of each addition. Introducing the MIB (Base+MIB) enhances PSNR by \textbf{+0.48 dB}. Including CA (Base+MIB+CA) further boosts performance by \textbf{+0.78 dB}. Adding all components (Base+MIB+CA+SA) significantly improves performance by \textbf{+1.28 dB}. This demonstrates the superiority of each part in AMIB. We further explore the effectiveness of CA and SA with visual description. In Figure~\ref{fig:amib_attn}(e), characterized by a brighter flare region, a higher CA score is attained compared to Figure~\ref{fig:amib_attn}(d), which has a darker flare region. Also, the SA highlights flare regions as depicted in Figure~\ref{fig:amib_attn}(f). This directs SFIM to emphasize flare-related features. In summary, the newly introduced AMIB adeptly integrates spatial and frequency features across all levels, directing the network's focus toward critical channels and spatial regions.

\paragraph{Optimal configuration of the multi-level architecture.}
\label{sec:exp_ablation_global}
As shown in Table~\ref{tab:role_assignment}, SFIM achieves the best result when placing SDBs at level 1 and FDBs at lower levels. The performance progressively decreases even with more parameters when SDBs are placed on more levels. Additionally, placing FDBs in all four levels performs worse than SFIM's configuration. This implies that SFIM effectively assigns roles to CNNs and FFT-based models by leveraging their inherent processing characteristics.

\begin{table}[t]
\centering
\begin{minipage}[t]{\textwidth}
\centering
\caption{Ablation study for the optimal configuration of the multi-level architecture on the UDC-SIT~\cite{ahn2024udc} dataset. The performance presented in this table corresponds to the SFIM trained up to phase 1. The number in parentheses denotes the designated levels for each block. Each column’s best and second-best scores are bold-faced and underlined, respectively.}
\label{tab:role_assignment}
\resizebox{0.6\columnwidth}{!}{
    \begin{tabular}{@{}c|c|cccc@{}}
    \toprule
    Configuration          & Param (M) & PSNR~$\uparrow$    & SSIM~$\uparrow$ \\ \midrule
                 FDB (1-4) &   23.29   & 29.10 & \underline{0.8742} \\         
    SDB ( 1 ) \& FDB (2-4) &   24.89   & \textbf{29.47} & \underline{0.8742} \\     
    SDB (1-2) \& FDB (3-4) &   31.63   & \underline{29.19} & \textbf{0.8744} \\     
    SDB (1-3) \& FDB ( 4 ) &   59.35   & 29.13 & 0.8696 \\
    SDB (1-4)              &   87.07   & 28.82 & 0.8674 \\ \bottomrule
    \end{tabular}
}
\end{minipage}%
\hfill
\centering
\vspace{1.0\baselineskip}
\begin{minipage}[t]{0.48\textwidth}
\centering
\caption{Ablation study on the embedding dimension of SFIM. The performance presented in this table corresponds to the SFIM model trained up to phase 1 on the UDC-SIT~\cite{ahn2024udc} dataset.}
\label{tab:exp_dimension}
\resizebox{0.85\textwidth}{!}{
    \begin{tabular}{@{}c|ccccc@{}}
    \toprule
    Dim. & Param (M) & PSNR~$\uparrow$    & SSIM~$\uparrow$ \\ \midrule
    24   &    6.72   & 29.04 & 0.8699 \\ 
    36   &   14.37   & 28.94 & 0.8708 \\     
    48   &   24.89   & 29.47 & 0.8742 \\ \bottomrule
    \end{tabular}
}
\end{minipage}%
\hfill
\begin{minipage}[t]{0.48\textwidth}
\centering
\caption{Comparing SFIM and ECFNet~\cite{zhu2022enhanced} using the UDC-SIT~\cite{ahn2024udc} dataset.}
\label{tab:ablation_levels}
\resizebox{0.8\textwidth}{!}{
    \begin{tabular}{@{}cc|cccc@{}}
    \toprule
     \# Levels & Models     & PSNR~$\uparrow$    & SSIM~$\uparrow$ \\ \midrule
    1        & SFIM   & 24.21  & 0.7349 \\     
             & ECFNet & \textbf{24.76}  & \textbf{0.7870} \\ \bottomrule
    2        & SFIM   & 26.29 & \textbf{0.8446} \\     
             & ECFNet & \textbf{26.32}  & 0.8131 \\ \bottomrule
    4        & SFIM   & \textbf{29.61} & \textbf{0.8758} \\     
             & ECFNet & 28.26  & 0.8491 \\ \bottomrule
    \end{tabular}
}
\end{minipage}
\end{table}

\begin{figure}[!t]
  \centering
   \includegraphics[width=1.0\linewidth]{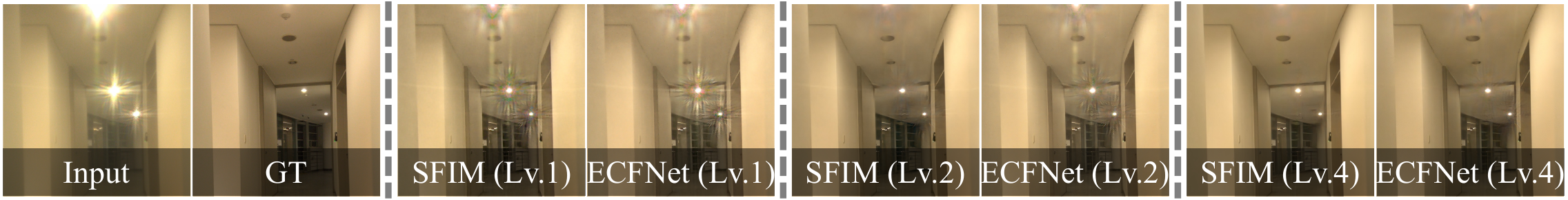}
   \vspace{-1.5\baselineskip}
   \caption{Comparison of restored images by SFIM and ECFNet trained with varying levels.}
   \label{fig:04_ablation_num_levels}
\end{figure}

\paragraph{Embedding dimension.}
\label{sec:exp_ablation_embed}
SFIM uses an embedding dimension 48 at level 1, as shown in Figure~\ref{fig:03_SFIM}. We gradually decrease the embedding dimension from 48 to 36 (SFIM-M) and then to 24 (SFIM-T) (Table~\ref{tab:exp_dimension}). Notably, even with an embedding dimension of 24 and fewer parameters (6.72 M), SFIM-T still exhibits superior performance to ECFNet and FFTformer (Table~\ref{tab:exp_quantitative_sfim} and Table~\ref{tab:exp_dimension}).

\paragraph{Effect of number of levels.} 
\label{sec:exp_ablation_level}
We explore the impact of incorporating spatial and frequency information in the multi-level architectures by gradually increasing the number of levels in SFIM and ECFNet~\cite{zhu2022enhanced}. ECFNet, top-ranked in the MIPI challenge~\cite{feng2022mipi}, uses CNNs at all levels. This comparison highlights the importance of utilizing both spatial and frequency domains in a multi-level architecture. As shown in Table~\ref{tab:ablation_levels}, SFIM outperforms ECFNet when increasing the number of levels. Figure~\ref{fig:04_ablation_num_levels} shows that both models handle blur and noise well at level 1 but struggle with flares. SFIM excels over ECFNet in flare restoration at levels 2 and 4, demonstrating the effectiveness of multi-level architectures that integrate spatial and frequency information to address multiple degradation patterns across extensive areas. CNNs in a multi-level architecture lack the receptive field to effectively manage flares.

%% file: sec/06_Conclusion.tex
\section{Conclusion}
\label{sec:conclusion}
In this paper, we study UDC image restoration both in the spatial and the frequency domains, introducing a novel approach that integrates spatial and frequency information in a multi-level architecture. We propose spatial and frequency interactive learning in the multi-level architecture (SFIM), which includes the spatial domain block (SDB), frequency domain blocks (FDB), and new attention-based multi-level integration block (AMIB). This design adeptly addresses local degradations, such as noise and blur, as well as global degradations, such as flares. The AMIB efficiently integrates information across all levels in the multi-level architecture, guiding the network to emphasize crucial areas. Experimental results on three UDC benchmarks indicate that SFIM is superior to the existing state-of-the-art image restoration models. 

%% file: sec/07_Acknowledgment.tex
\section*{Acknowledgment}

This work was supported in part by the National Research Foundation of Korea (NRF) grant (No. RS-2023-00222663, Center for Optimizing Hyperscale AI Models and Platforms), by the Institute for Information and Communications Technology Promotion (IITP) grant (No. 2018-0-00581, CUDA Programming Environment for FPGA Clusters), by the BK21 Plus programs for Innovative Data Science Talent Education Program (Dept. of Data Science, SNU, No. 5199990914569) and BK21 FOUR Intelligence Computing (Dept. of Computer Science and Engineering, SNU, No. 4199990214639) through the NRF, all funded by the Ministry of Science and ICT (MSIT) of Korea. This work was also supported in part by the Samsung Display Co., Ltd. ICT at Seoul National University provided research facilities for this study.

\clearpage

%% file: sec/09_Appendix.tex
\appendix 

\setcounter{figure}{0}  
\setcounter{table}{0} 
\counterwithin{figure}{section}
\counterwithin{table}{section}

\section*{\centering\textbf{Appendix}}
In this Appendix, Section~\ref{sec:appendix_background} provides background information on UDC degradation and describes our model architecture's rationale in light of the UDC dataset's characteristics. Section~\ref{sec:appendix_ablation} discusses the importance of integrating frequency information through an ablation study on the FFT loss terms. Section~\ref{sec:appendix_evaluation} presents additional experiments. Section~\ref{sec:appendix_training_details} provides detailed information on the training of SFIM. Finally, Section~\ref{sec:limitations} outlines the limitations of this work.

\section{Background and Rationale}
\label{sec:appendix_background}

\begin{figure}[b]
  \centering
  \includegraphics[width=0.35\textwidth]{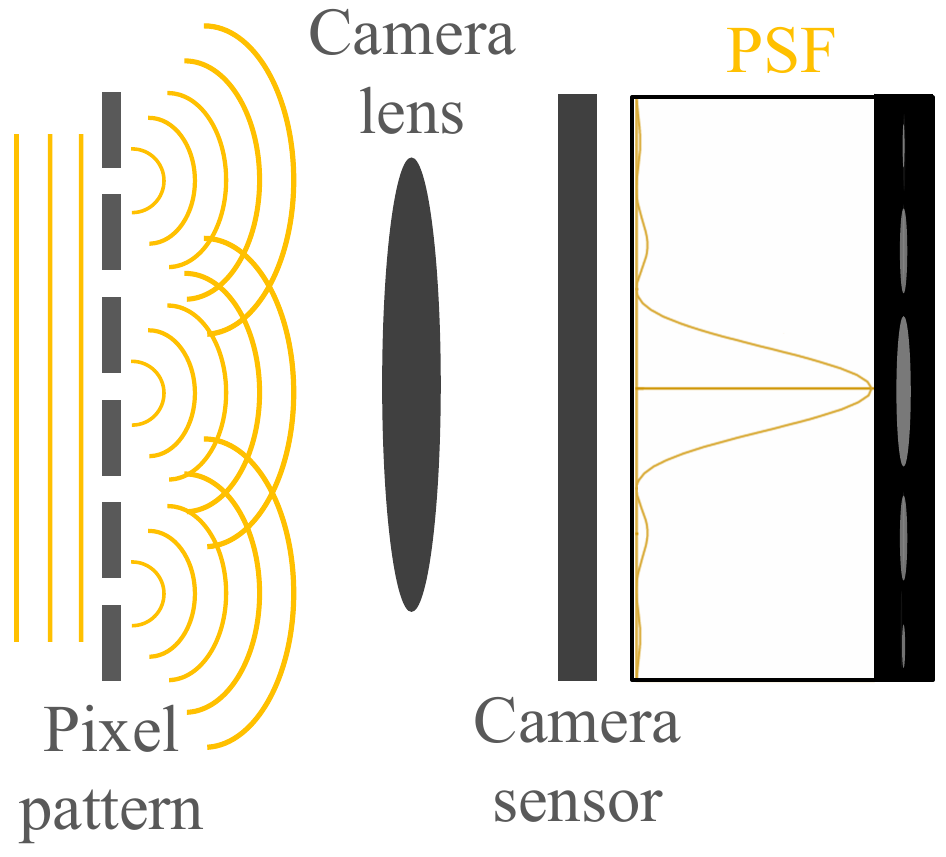}
  \caption{The PSF is determined by the pixel arrangement on the UDC area.}
  \label{fig:appendix_diffraction}
\end{figure}

In the UDC setup, the camera is placed beneath the display. The display pixels act as slits, causing diffraction in captured images (Figure~\ref{fig:appendix_diffraction}). Different pixel designs produce unique diffraction and flare patterns. As a result, the SYNTH~\cite{feng2021removing} dataset (ZTE Axon 20~\cite{zteAxon20}'s degradation) and UDC-SIT~\cite{ahn2024udc} dataset (Samsung Galaxy Z-Fold 3~\cite{samsungGalaxyZFold3}'s degradation) show distinct flare shape. Figure~\ref{fig:01_frequency_analysis}(f) and (g) describe the distinct degradation pattern between them. The light source's position also affects the flare shape, observed solely in the UDC-SIT dataset. 

Among the three UDC datasets, such as T-OLED~\cite{zhou2021image}, SYNTH~\cite{feng2021removing}, and UDC-SIT~\cite{ahn2024udc}, we focus on the UDC-SIT dataset, which is the only one with ultra-high definition (UHD) resolution. In higher-resolution images captured by modern smartphones, flare affects many pixels. Thus, CNNs with multi-level architecture still lack sufficient receptive fields, even at lower levels. Handling \textit{spatially variant flares} with invariant CNN kernels is also challenging. Beyond flares, the UDC-SIT dataset also suffers from severe local degradation, such as noise and blur. We focus on the real-world dataset with the aforementioned characteristics. SFIM uses CNNs at level 1 to manage local degradation, while FFT-based models at lower levels capture frequency priors for global degradation. The proposed AMIB effectively integrates both spatial and frequency features across all levels. 

Table~\ref{tab:datasets_comparison} offers an overview of the dataset characteristics, followed by detailed information below.

\paragraph{T-OLED.}
Zhou et al.~\cite{zhou2021image} collect the paired UDC images from a Monitor Camera Imaging System (MCIS). This system captures paired images in a controlled environment. Specifically, images are displayed on a monitor and captured in two setups, one with a T-OLED or P-OLED display in front of the camera for degraded images and one without the display for ground-truth images. Despite their pioneering work, flares are almost absent due to the limited dynamic range of the monitor. Moreover, they lack actual UDC degradations, as depicted in Figure~\ref{fig:01_frequency_analysis} in the main body of the paper. The dataset comprises 300 pairs of images.

\paragraph{SYNTH.}
Feng et al.~\cite{feng2021removing} improve the UDC dataset. They measure the point spread function (PSF) of ZTE Axon 20~\cite{zteAxon20} and convolve the PSF with high dynamic range (HDR) images from the HDRI Haven dataset~\cite{hdrihaven}. Consequently, the SYNTH dataset exhibits flare patterns. However, it has limitations, including the absence of noise and overly regular flare occurrences~\cite{ahn2024udc}. The dataset comprises 2,376 images.

\paragraph{UDC-SIT.}
Ahn et al.~\cite{ahn2024udc} propose a real-world UDC dataset using their proposed image-capturing system. They cut the UDC display panel of Samsung Galaxy Z-Fold 3~\cite{samsungGalaxyZFold3} and attach it to a lid. They acquire the paired images by opening and closing the lid attached to the standard camera~\cite{samsungGalaxyNote10}. Alignment of paired images is achieved through discrete wavelet Fourier transform (DFT), with an accuracy of 97.26\% measured by the percentage of correct keypoints (PCK)~\cite{feng2023generating}. The UDC-SIT dataset faithfully represents real UDC degradations, particularly noteworthy for its realistic flares, which occasionally cover the entire image and exhibit spatial variance in an image. Additionally, this dataset boasts a UHD resolution and consists of 2,340 images.

\begin{table}[t]
  \caption{Comparison of the UDC datasets.}
  \label{tab:datasets_comparison}
  \centering
  \resizebox{1.0\linewidth}{!}{%
  \begin{tabular}{lcccccc}
    \toprule
    Dataset                            & Flare existence & Resolution & \#Training & \#Validation & \#Test & \#Total \\
    \midrule
    T-OLED~\cite{zhou2021image} &           & [1024, 2048, 3] & 240   & 30  & 30  & 300 \\
    SYNTH~\cite{feng2021removing}      & \ding{52} & [800, 800, 3]   & 2,016 &  -  & 360 & 2,376 \\
    UDC-SIT~\cite{ahn2024udc}          & \ding{52} & [1792, 1280, 4] or [3584, 2560, 3] & 1,864 & 238 & 238 & 2,340 \\
    \bottomrule
  \end{tabular}
  }
\end{table}

\section{Ablation Study}
\label{sec:appendix_ablation}
In this section, we present further ablation studies focusing on the FFT loss terms.

\subsection{Inclusion of the FFT Loss}
As discussed in Section~\ref{sec:03_Motivation} and Section~\ref{sec:exp_ablation} in the main body of the paper as well as Section~\ref{sec:appendix_background}, integrating local and global information is crucial, particularly when handling the UDC-SIT dataset. Furthermore, using the loss in both spatial (e.g., the Charbonnier loss and the SSIM loss) and frequency (e.g., the FFT loss; $\mathcal{L}_{FFT, l}$) domains is also crucial.

As illustrated in Section~\ref{sec:login_loss} in the main body of the paper, we use two FFT loss terms: the amplitude term $\mathcal{L}_{\mathcal{A}, l}$ and the phase term $\mathcal{L}_{\phi, l}$, as formulated in Equation~\ref{eq:fft_amp} in the main body of the paper. ECFNet, another multi-level architecture featuring CNNs at each level, also use the FFT loss, as elaborated below:

\begin{equation}
  \label{eq:loss_sfim_ecfnet}
  \mathcal{L}_{l} = \mathcal{L}_{Char, l} + \lambda_1 \cdot \mathcal{L}_{FFT, l} \; \mbox{and}
\end{equation}

\begin{equation}
  \label{eq:loss_fft_ecfnet}
  \mathcal{L}_{FFT, l} = \sum\limits_{u=0}^{M_l - 1}\sum\limits_{v=0}^{N_l - 1} || \mathcal{F}_{R}(u,v) - \mathcal{F}_{G}(u,v) ||_1,
\end{equation}

where $(M_l, N_l)$ is size of the image at level $l$, $(u,v)$ is a point in the frequency domain, $\mathcal{F}_{R}(u,v)$ and $\mathcal{F}_{G}(u,v)$ denote the FFT of the restored and ground-truth images, respectively. 

As described in Table~\ref{tab:sup_01_ablation_fft_loss}, the inclusion of the FFT loss ($\mathcal{L}_{FFT, l}$) yields superior performance in both SFIM and ECFNet. The quantitative differences between experiments with and without $\mathcal{L}_{FFT, l}$ are more pronounced in the UDC-SIT and SYNTH datasets compared to the T-OLED dataset. This result underscores the importance of discerning \textit{intrinsic frequency priors that imply the presence of the flares} in addressing flare-related challenges.

Specifically, SFIM exhibits PSNR gaps between experiments with and without $\mathcal{L}_{FFT, l}$ are \textbf{1.50 dB} and \textbf{2.90 dB} for the UDC-SIT and SYNTH datasets, respectively. Conversely, the observed PSNR gaps in ECFNet are \textbf{0.71 dB} and \textbf{3.60 dB} for the UDC-SIT and SYNTH datasets, respectively. This discrepancy is due to flares in the UDC-SIT and SYNTH datasets. The difference between the two models is more noticeable in the UDC-SIT dataset, given its UHD resolution and \textit{spatially variant flare}, requiring more substantial global information than the SYNTH dataset. SFIM's performance on the UDC-SIT dataset without $\mathcal{L}_{FFT, l}$ is similar to that of ECFNet with $\mathcal{L}_{FFT, l}$. This finding emphasizes SFIM's ability to effectively learn \textit{intrinsic frequency priors} without relying on FFT loss terms, as the model efficiently integrates spatial and global information within its optimal multi-level architecture configuration. However, in the T-OLED dataset, this PSNR gap reduces to \textbf{0.08 dB} and \textbf{0.17 dB} for SFIM and ECFNet, respectively. This decrease is because flares are absent in the T-OLED dataset. 

Remember that the spectral amplitude difference between degraded and ground-truth images exhibits distinct long and thin peaks in the presence of flares, as depicted in Figure~\ref{fig:01_frequency_analysis} in the main body of the paper. It is evident that learning in the frequency domain significantly contributes to restoring UDC-degraded images containing flares.

\begin{table}[t!]
  \small
  \vspace{-0.5\baselineskip}
  \caption{The restoration performance of SFIM and ECFNet~\cite{zhu2022enhanced} on the three datasets both with and without the FFT loss ($\mathcal{L}_{FFT}$). The term \textit{Flare} denotes the presence of flares in the dataset.} 
  \label{tab:sup_01_ablation_fft_loss}
  \centering
  \resizebox{0.99\linewidth}{!}{
  \begin{tabular}{>{\centering\arraybackslash}m{2.0cm}|>{\centering\arraybackslash}m{2.0cm}|>{\centering\arraybackslash}m{1.8cm}|>{\centering\arraybackslash}m{1.8cm}|>{\centering\arraybackslash}m{1.8cm}|>{\centering\arraybackslash}m{1.8cm}|>{\centering\arraybackslash}m{1.7cm}}
    \toprule
    \multirow{2}{*}{Dataset} & \multirow{2}{*}{Weights} & \multicolumn{2}{c|}{SFIM} & \multicolumn{2}{c|}{ECFNet~\cite{zhu2022enhanced}} & \multirow{2}{*}{Flare} \\
    \cline{3-6}
     &  & w/ $\mathcal{L}_{FFT,l}$ & w/o $\mathcal{L}_{FFT,l}$ & w/ $\mathcal{L}_{FFT,l}$ & w/o $\mathcal{L}_{FFT,l}$ & \\ 
    \midrule
    \multirow{2}{*}{UDC-SIT~\cite{ahn2024udc}}
            & PSNR $\uparrow$ & \textbf{29.61} & 28.11 & \textbf{28.26} & 27.55 & \multirow{2}{*}{\ding{52}} \\
            & SSIM $\uparrow$ & \textbf{0.8758} & 0.8592 & 0.8491 & \textbf{0.8559} &  \\
    \midrule
    \multirow{2}{*}{SYNTH~\cite{feng2021removing}} 
            & PSNR $\uparrow$ & \textbf{49.77} & 46.87 & \textbf{47.75} & 44.15 & \multirow{2}{*}{\ding{52}} \\
            & SSIM $\uparrow$ & \textbf{0.9940} & 0.9917 & \textbf{0.9915} & 0.9864 &  \\
    \midrule
    \multirow{2}{*}{T-OLED~\cite{zhou2021image}} 
            & PSNR $\uparrow$ & \textbf{37.62} & 37.54 & \textbf{38.19} & 38.02 & \multirow{2}{*}{} \\
            & SSIM $\uparrow$ & \textbf{0.9530} & 0.9520 & \textbf{0.9550} & 0.9529 &  \\
    \bottomrule
  \end{tabular}
  }
\end{table}

\subsection{Amplitude and Phase Components of FFT Loss}
FFT amplitude represents a signal's magnitude in the frequency domain. It indicates the intensity of each frequency component, with a higher amplitude indicating a stronger frequency component. For example, the amplitude is associated with the image's brightness in low-light enhancement tasks~\cite{wang2023fourllie}. On the other hand, the phase component is essential for preserving vital structural information~\cite{wang2023spatial}. Thus, both amplitude and phase information are essential in addressing UDC degradation, such as low transmittance and flares.

\begin{figure}[!b]
  \centering
  \begin{minipage}{0.6\linewidth}
  \centering
  \includegraphics[width=1.0\linewidth]{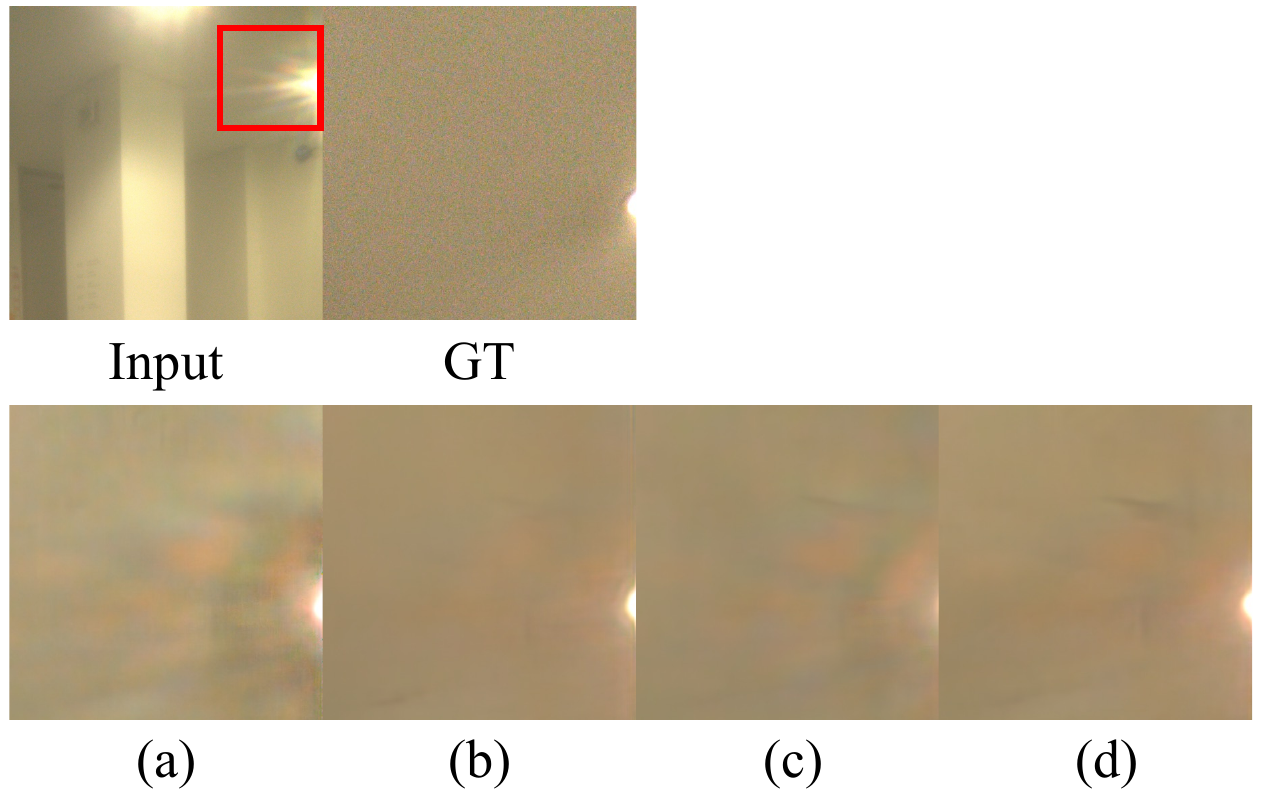}
   \end{minipage}{}
   \vspace{-0.5\baselineskip}
   \caption{Ablation study for the FFT terms in the loss function. (a) Without the amplitude and phase terms. (b) With the amplitude term only. (c) With the phase term only. (d) With both the amplitude and phase terms.}
   \label{fig:sup_01_fft_loss_terms}
\end{figure}

As shown in Table~\ref{tab:sup_01_ablation_fft_loss_terms}, the best performance in SFIM is observed when using a loss function that incorporates both the amplitude and phase terms of FFT. The absence of FFT-related terms in the loss function results in the worst performance, with a significant PSNR decrease of \textbf{1.50 dB} compared to the model using both terms. Notably, considering either the amplitude or phase term alone results in a considerable performance improvement. Nevertheless, the optimal performance is achieved when both terms are taken into account. The visual representation in Figure~\ref{fig:sup_01_fft_loss_terms} demonstrates the importance of incorporating FFT terms in the loss function for effectively restoring the flare. 

\begin{table}[t]
  \small
  \vspace{-0.5\baselineskip}
  \caption{The restoration performance of SFIM on the UDC-SIT~\cite{ahn2024udc} dataset with respect to the amplitude and phase components of FFT within the loss function.}
  \label{tab:sup_01_ablation_fft_loss_terms}
  \centering
  \resizebox{0.85\linewidth}{!}{
  \begin{tabular}{>{\centering\arraybackslash}m{2.0cm}|>{\centering\arraybackslash}m{1.8cm}|>{\centering\arraybackslash}m{1.8cm}|>{\centering\arraybackslash}m{1.8cm}|>{\centering\arraybackslash}m{1.8cm}}
    \toprule
    \multirow{2}{*}{Metric} & Amplitude & Amplitude & Phase & \multirow{2}{*}{None} \\
     & \& phase & only & only &  \\ 
    \midrule
            PSNR $\uparrow$                & \textbf{29.61}   & 29.51   & 29.30 & 28.11 \\
            SSIM $\uparrow$                & \textbf{0.8758}  & 0.8739  & 0.8742 & 0.8592  \\
    \bottomrule
  \end{tabular}
  }
\end{table}

\section{Additional Evaluation}
\label{sec:appendix_evaluation}
In this section, we decribe additional quantitative and qualitative results.

\subsection{Quantitative Results}
\label{sec:appendix_quantitative}
In this section, we provide quantitative result on the T-OLED dataset~\cite{zhou2021image}. As shown in Table~\ref{tab:exp_toled}, FSI~\cite{liu2023fsi}, which integrates spatial and frequency domain information with a dual-stream network, exhibits superior performance compared to SFIM on the T-OLED dataset. However, our proposed SFIM outperforms FSI with a large margin (\textbf{3.63 dB}) in the SYNTH dataset containing flares as described in Table~\ref{tab:exp_quantitative_sfim} in the main body of the paper. 

\begin{table}[h]
  \small
  \caption{Restoration performance for the T-OLED dataset~\cite{zhou2021image}. The \textit{Input} column has the PSNR and SSIM values between the paired degraded and ground-truth images. Each row’s best and second-best scores are bold-faced and underlined, respectively.} 
  \label{tab:exp_toled}
  \centering
  \resizebox{1.0\linewidth}{!}{
  \begin{tabular}{lcccccccccc}
    \toprule
                  & Input & UNet~\cite{zhou2021image} & IPIUer~\cite{zhou2020udc} & BAIDU~\cite{zhou2020udc} & DAGF~\cite{sundar2020deep} & BNUDC~\cite{koh2022bnudc} & ECFNet~\cite{zhu2022enhanced} & PDCRN~\cite{panikkasseril2020transform} & FSI~\cite{liu2023fsi} & SFIM (Ours) \\
    \midrule
    PSNR~$\uparrow$ & 28.44          & 36.71   & 38.18  & \underline{38.23}  & 36.91 & 38.26 & 38.19   & 37.83 & \textbf{38.60} & 37.62 \\
    SSIM~$\uparrow$ & 0.8221         & 0.971   & \underline{0.979}  & \textbf{0.980}  & 0.973 & \textbf{0.980} & 0.955  & 0.978 & \textbf{0.980} & 0.953 \\
    \bottomrule
  \end{tabular}
  }
\end{table}

\subsection{Qualitative Results}
\label{sec:appendix_qualitative}
\begin{figure}[!b]
  \centering
  \begin{minipage}{0.60\linewidth}
  \centering
   \includegraphics[width=1.0\linewidth]{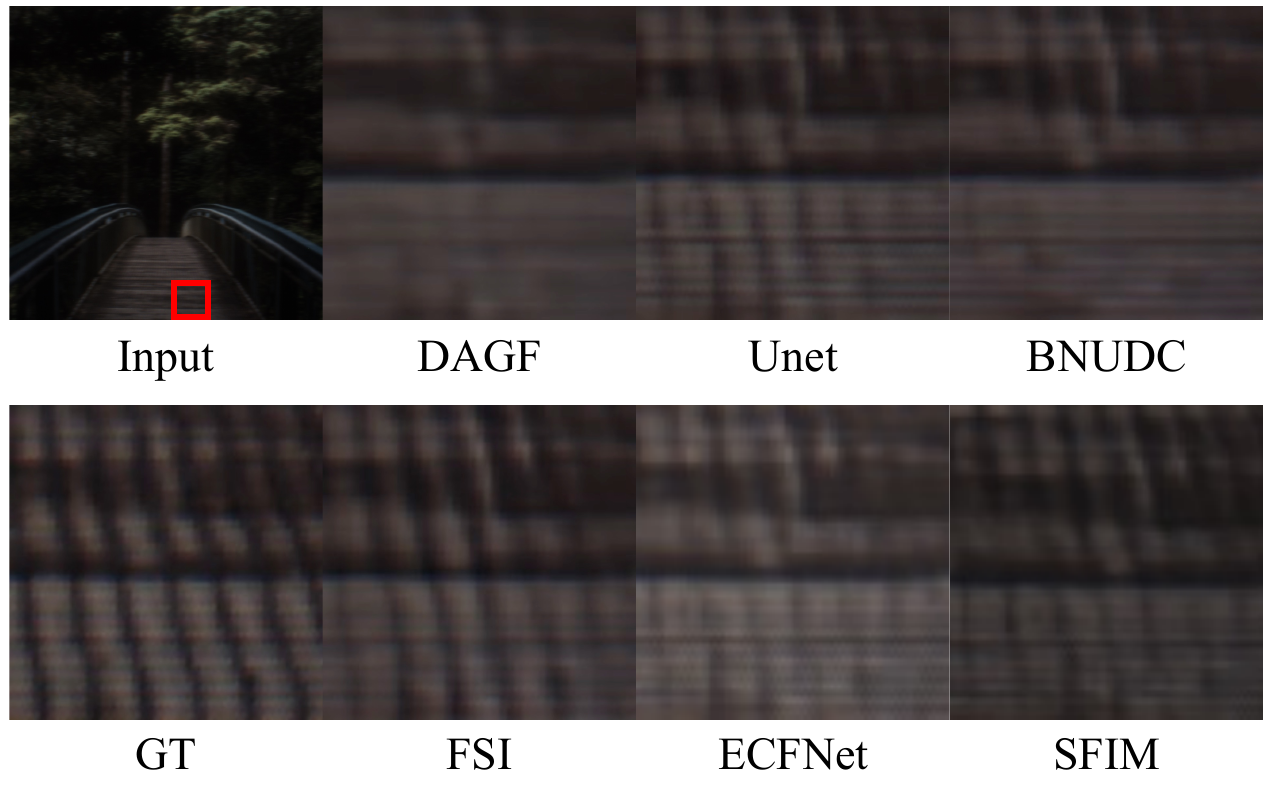}
   \caption{T-OLED~\cite{zhou2021image} images restored by the models.}
   \label{fig:sup_02_qualitative_4}
   \end{minipage}{}
\end{figure}

This section provides additional qualitative results of SFIM and other models on the three datasets. In the T-OLED~\cite{zhou2021image} dataset, SFIM proficiently reconstructs the intricate wooden texture on the bridge as illustrated in Figure~\ref{fig:sup_02_qualitative_4}.

\begin{figure}[t!]
  \centering
  \begin{minipage}{0.90\linewidth}
  \centering
   \includegraphics[width=1.0\linewidth]{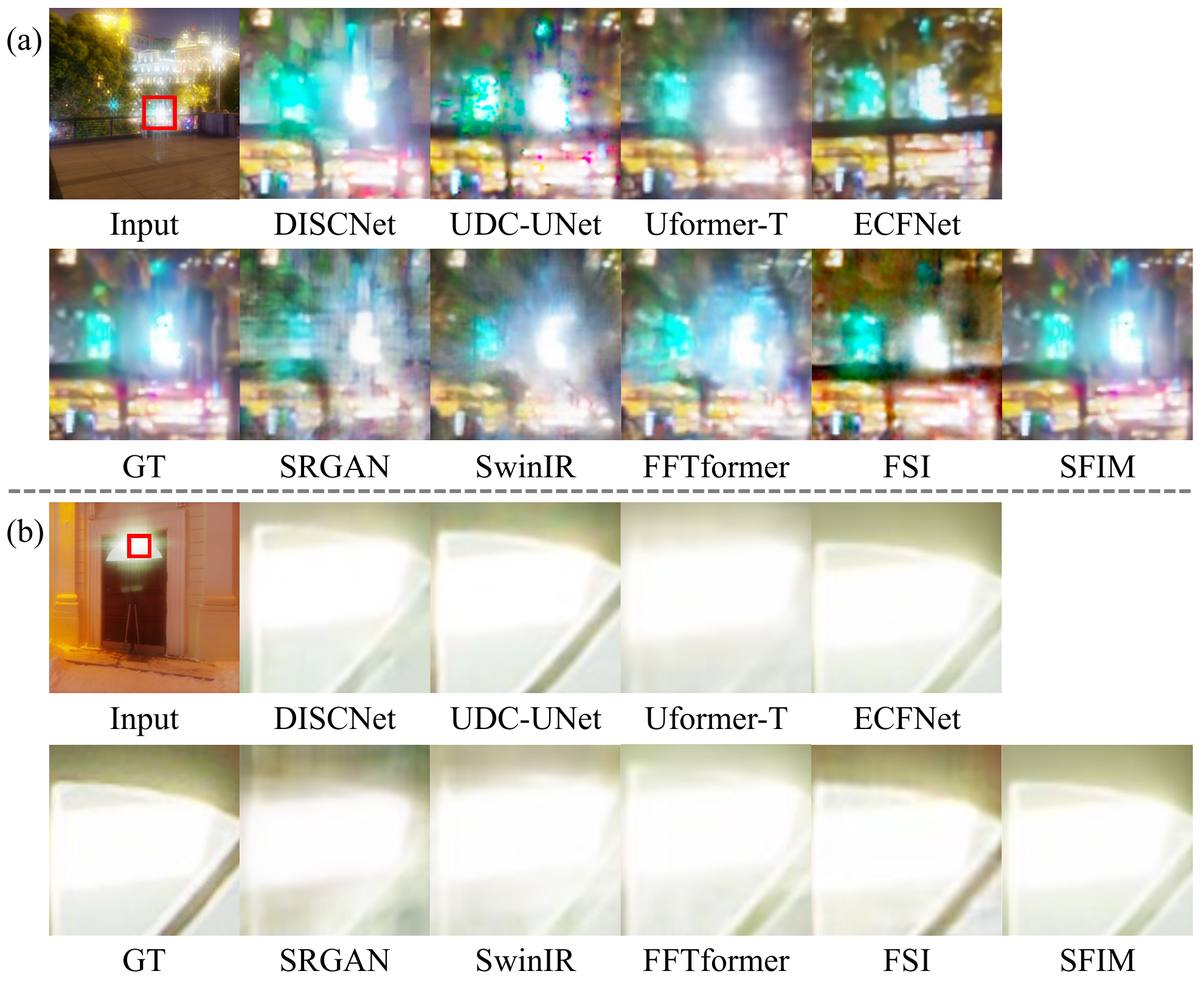}
   \caption{SYNTH~\cite{feng2021removing} images restored by the models.}
   \label{fig:sup_02_qualitative_3}
   \end{minipage}{}
\end{figure}

Figure~\ref{fig:sup_02_qualitative_3} shows the superiority of SFIM for the SYNTH~\cite{feng2021removing} dataset. One interesting observation is that although ECFNet seems to have restored the handle of the fence and the leaves most effectively, SFIM has achieved the closest similarity to the ground-truth image as illustrated in Figure~\ref{fig:sup_02_qualitative_3}(a). It is excessive restoration to unconditionally connect the fence, which is not connected in the ground-truth, solely based on local information. ECFNet is a CNN-only model solely in the spatial domain, while SFIM integrates local and global information both in spatial and frequency domains. This excessive restoration still occurs in FSIE even though FSI integrates both spatial and frequency information. Furthermore, SFIM more successfully recovers the details of the light source (lamp) compared to other models, as shown in Figure~\ref{fig:sup_02_qualitative_3}(b).

\begin{figure}
  \centering
  \begin{minipage}{0.79\linewidth}
  \centering
   \includegraphics[width=1.0\linewidth]{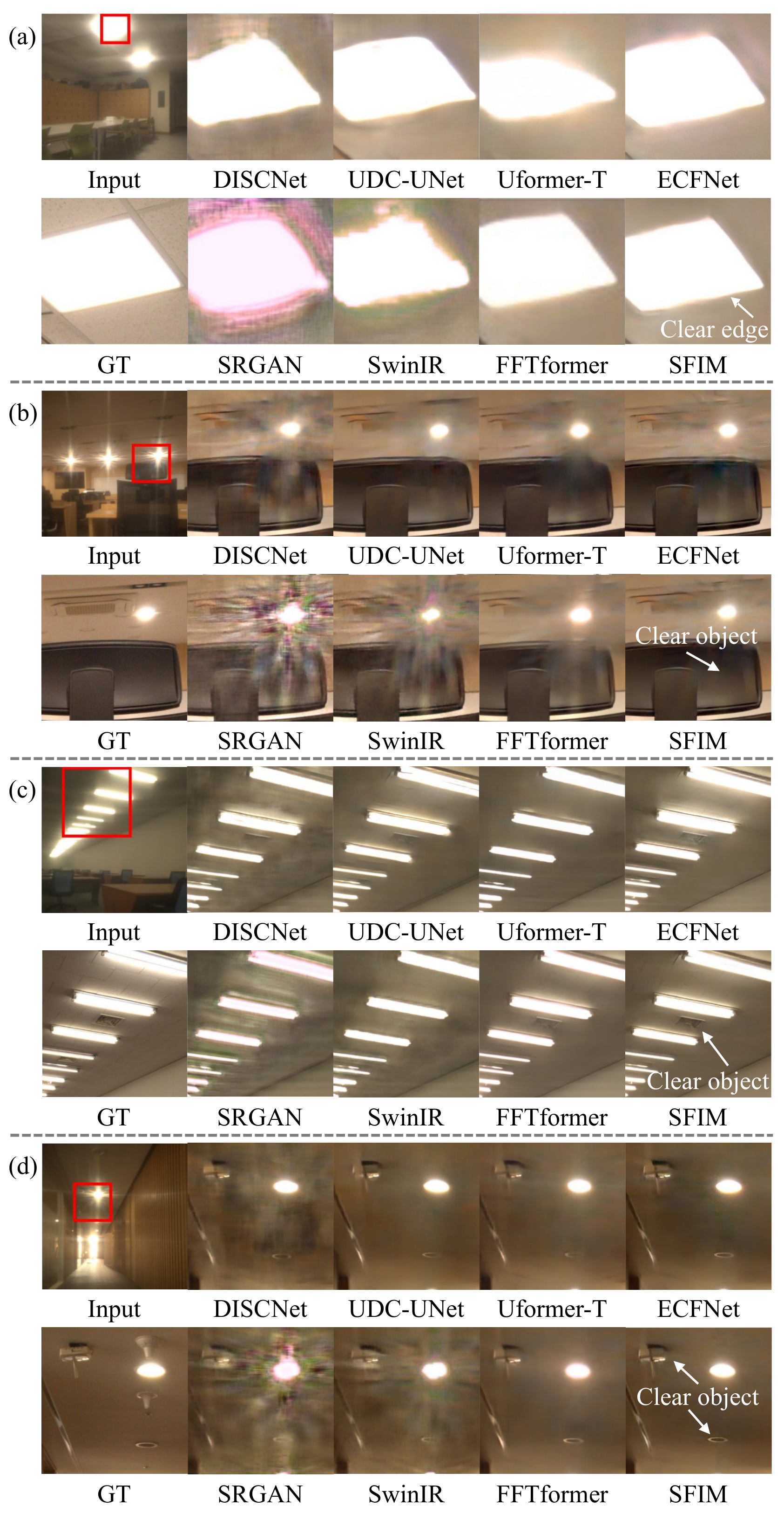}
   \caption{UDC-SIT~\cite{ahn2024udc} images restored by the models.}
   \vspace{-0.5\baselineskip}
   \label{fig:sup_02_qualitative_1}
   \end{minipage}{}
\end{figure}

\begin{figure}
  \centering
  \begin{minipage}{0.79\linewidth}
  \centering
   \includegraphics[width=1.0\linewidth]{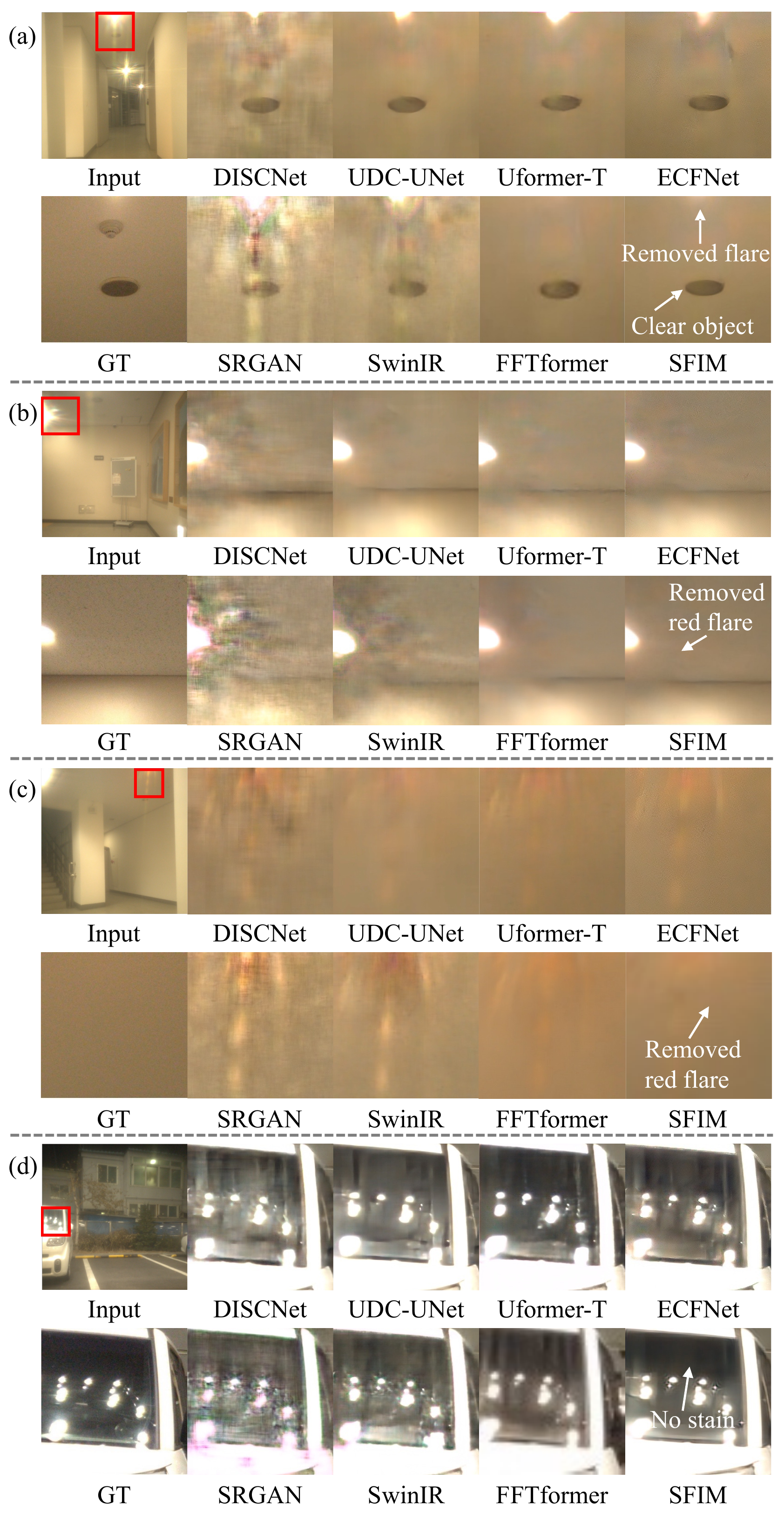}
   \caption{UDC-SIT~\cite{ahn2024udc} images restored by the models.}
   \vspace{-0.5\baselineskip}
   \label{fig:sup_02_qualitative_2}
   \end{minipage}{}
\end{figure}

In the UDC-SIT dataset~\cite{ahn2024udc}, Figure~\ref{fig:sup_02_qualitative_1} and Figure~\ref{fig:sup_02_qualitative_2} effectively demonstrate SFIM's superior performance to the other models. Sometimes, other models encounter difficulties in restoring light sources with precise edges; however, SFIM showcases precise edges around the light sources, as demonstrated in Figure~\ref{fig:sup_02_qualitative_1}(a). Moreover, SFIM demonstrates strong proficiency in the intricate task of restoring obscured objects as depicted in Figure~\ref{fig:sup_02_qualitative_1}(b)-(d) and Figure~\ref{fig:sup_02_qualitative_2}(a). Also, the restoration of flares caused by light sources positioned at the end or beyond an image, where no explicit light source is present in the ground-truth image, presents a challenging task. SFIM surpasses other models in effectively restoring these flares as shown in Figure~\ref{fig:sup_02_qualitative_2}(a)-(c). Finally, SFIM tends to generate either no stains while removing flares in the UDC-degraded images, unlike other models as shown in Figure~\ref{fig:sup_02_qualitative_2}(d). These qualitative findings highlight SFIM's superior performance to existing models.

\section{Training Details}
\label{sec:appendix_training_details}

\subsection{Progressive Training in SFIM}
SFIM is initially trained on smaller images, gradually progressing to larger images in later phases. The detailed configuration is described in Table~\ref{tab:sup_02_progressive_learning}. 

\begin{table}[hb!]
\centering
\caption{Progressive learning strategy for the three datasets. LR means the initial learning rate. The learning rate is decayed with a cosine annealing schedule for the given epochs, where $\eta_{min} = 1 \times 10^{-7}$ and $\eta_{max} = 2 \times 10^{-4}$. \label{tab:sup_02_progressive_learning}} 
\resizebox{0.68\columnwidth}{!}{
\begin{tabular}{c|ccccc}
\toprule
Dataset                                        & Phase & LR       & Epochs & Patch size & Batch size \\ \midrule
\multirow{2}{*}{UDC-SIT~\cite{ahn2024udc}}     &   1   & 0.0002   & 1000   & 512        & 56 \\ 
                                               &   2   & 0.00001  & 150    & 768        & 28 \\ \midrule
\multirow{3}{*}{SYNTH~\cite{feng2021removing}} &   1   & 0.0002   & 1000   & 256        & 16 \\                       
                                               &   2   & 0.00001  & 300    & 512        & 16 \\
                                               &   3   & 0.000008 & 150    & 800        & 16 \\ \midrule
\multirow{3}{*}{T-OLED~\cite{zhou2021image}}   &   1   & 0.0002   & 4000   & 256        & 16 \\
                                               &   2   & 0.00001  & 1200   & 512        & 16 \\
                                               &   3   & 0.000008 & 600    & 800        & 16 \\ \bottomrule
\end{tabular}
}
\end{table}

\subsection{Details of Training the DNN Models}
This section provides detailed information about the DNN models used in the evaluation.

\subsubsection{System Configuration.}
We use a 10-node GPU cluster to train the benchmark models. Each node has four NVIDIA GeForce RTX 3090 GPUs. Table~\ref{tab:system_conf} shows the details of the system configuration and software used.

\begin{table}[h]
    \caption{System configuration of a node in the 10-node GPU cluster used for training.}
    \centering
    \resizebox{0.6\linewidth}{!}{
        \begin{tabular}{c|l}
        \toprule
            Motherboard & Supermicro H12DSG-O-CPU \\ \hline
            CPU & 2 $\times$ AMD EPYC 7502 32-Core Processor \\ \hline
            Memory & 8 $\times$ 64GB DDR4 DIMM \\ \hline
            GPU & 4 $\times$ NVIDIA Gefore RTX 3090 \\ \hline
            OS & Ubuntu 20.04.6 (kernel 5.4.0-100) \\ \hline
            GPU Driver & 520.61.05 \\ \hline
            CUDA Version & 11.7 \\ \hline
            PyTorch Version & 2.0.1 \\ \bottomrule
        \end{tabular}
    }
    \label{tab:system_conf}
\end{table}

\subsubsection{Training the Models.}
While we predominantly rely on the original authors' code for the benchmark models, we introduce some modifications to their code. We implement these adjustments, adhering to the specific modifications outlined in Ahn \textit{et al.}'s appendix~\cite{ahn2024udc}. As SwinIR~\cite{liang2021swinir} and FFTformer~\cite{kong2023efficient} are not included in their models evaluated, we train these models using a similar approach to the other models for modifications.

For SwinIR and FFTformer, we increase the models' input channel size to four to match the channel size of UDC-SIT. The models are trained using PyTorch Distributed Data Parallel (DDP)~\cite{pytorchddp} to accelerate training. The training specifications of SwinIR and FFTformer for the UDC-SIT dataset are as follows:

\begin{itemize}
\item {\bf SwinIR.}
We reduce the embedding dimension from 96 to 84 to overcome the GPU memory size limitation. Additionally, we decrease the number of Swin Transformer Layers (STL) within each Residual Swin Transformer Block (RSTB) from 6 to 4. Moreover, we adjust the number of heads to 4 for all STLs.

\item {\bf FFTformer.}
We reduce the embedding dimension from 48 to 24 to overcome the GPU memory size limitation. 
\end{itemize}

\section{Limitations} 
\label{sec:limitations}
Our focus on integrating both spatial and frequency domain information aligns with the characteristics of flares in UDC images, leading to superior performance compared to state-of-the-art methods. However, SFIM's effectiveness is less pronounced on the T-OLED dataset, where flares are absent. It is important to note that neither the T-OLED nor P-OLED datasets exhibit actual UDC degradation, such as flares.